\documentclass{article}
\usepackage[sort,compress]{cite}

\usepackage{geometry}
\usepackage[affil-it]{authblk}
\usepackage{url}
\usepackage{algorithmic}
\usepackage{amsfonts, amsmath, amssymb, amsopn, mathtools, bm, dsfont}
\usepackage{amsthm}
\newtheorem{theorem}{Theorem}[section]

\newtheorem{lemma}[theorem]{Lemma}
\newtheorem{corollary}[theorem]{Corollary}
\newtheorem{proposition}[theorem]{Proposition}
\newtheorem{remark}[theorem]{Remark}

\usepackage{booktabs}
\usepackage{color}
\usepackage{epstopdf}
\usepackage{float}
\usepackage{graphicx}
\usepackage{interval}
\usepackage{pgfplots}
\usepackage{subcaption}
\usepackage{wrapfig}
\usepackage{xargs}                      
\usepackage[colorinlistoftodos,prependcaption,textsize=tiny]{todonotes}
\usepackage{hyperref}

\newcommand{\changes}[1]{#1}

\newcommand{\oaival}[2]{\interval[open left]{#1}{#2}}
\newcommand{\aoival}[2]{\interval[open right]{#1}{#2}}
\newcommand{\R}{\mathbb{R}}
\newcommand{\N}{\mathbb{N}}
\newcommand{\cD}{\mathcal{D}}
\newcommand{\cN}{\mathcal{N}}
\newcommand{\cA}{\mathcal{A}}
\newcommand{\cB}{\mathcal{B}}
\newcommand{\cE}{\mathcal{E}}
\newcommand{\cS}{\mathcal{S}}
\newcommand{\cR}{\mathcal{R}}

\renewcommand{\epsilon}{\varepsilon}
\newcommand{\aaa}{\bm a}
\newcommand{\e}{\bm e}
\newcommand{\cc}{\bm c}

\newcommand{\sss}{\bm s}
\newcommand{\ttt}{\bm t}
\newcommand{\vv}{\bm v}
\newcommand{\w}{\bm w}
\newcommand{\xx}{\bm x} 
\newcommand{\y}{\bm y}
\newcommand{\z}{\bm z}
\newcommand{\A}{\bm A}
\newcommand{\I}{\bm I}

\newcommand{\M}{\bm M}
\newcommand{\U}{\bm U}
\newcommand{\V}{\bm V}

\newcommand{\RRR}{\bm R}
\newcommand{\SSS}{\bm S}
\newcommand{\T}{\bm T}

\newcommand{\sigmaa}{\bm \sigma}
\newcommand{\Sigmaa}{\bm \Sigma}
\newcommand{\thetaa}{\bm \theta}
\newcommand{\Thetaa}{\bm \Theta}

\newcommand{\Lambdaa}{\bm \Lambda}
\newcommand{\1}{\mathbf{1}}
\newcommand{\scp}[2]{\left\langle{#1},{#2}\right\rangle}
\newcommand{\mleq}{\preccurlyeq}
\newcommand{\mle}{\prec}
\newcommand{\mgeq}{\succcurlyeq}
\newcommand{\mge}{\succ}
\DeclarePairedDelimiter{\norm}{\lVert}{\rVert}

\DeclarePairedDelimiter{\abs}{\lvert}{\rvert}
\DeclareMathOperator*{\argmin}{argmin}
\DeclareMathOperator*{\arginf}{arginf}

\DeclareMathOperator{\diag}{diag}
\DeclareMathOperator{\Eig}{Eig}

\DeclareMathOperator{\Var}{Var}
\DeclareMathOperator{\trace}{trace}
\DeclareMathOperator*{\E}{\mathbb{E}}

\ifpdf
  \DeclareGraphicsExtensions{.eps,.pdf,.png,.jpg}
\else
  \DeclareGraphicsExtensions{.eps}
\fi

\begin{document}

\newcommand*{\email}[1]{%
    \normalsize\href{mailto:#1}{\texttt{#1}}\par%
    }
\markboth{C. Brauer, N. Breustedt, T. de Wolff, D. A. Lorenz}{Learning Variational Models with Unrolling and Bilevel Optimization}

\title{Learning Variational Models with Unrolling and Bilevel Optimization\footnote{This work was funded by the Federal Ministry of Education and Research (BMBF) under grant number 05M2020 (project MaGriDo)}}

\author[1]{Christoph Brauer\thanks{\email{christoph.brauer@dlr.de}}}

\affil[1]{Institute of Lightweight Systems, German Aerospace Center, Ottenbecker Damm 12\\
21684 Stade, Germany}

\author[2]{Niklas Breustedt\thanks{\email{n.breustedt@tu-braunschweig.de}}}

\affil[2]{Institute of Analysis and Algebra, Technical University of Braunschweig, Universit\"atsplatz 2\\
38106 Braunschweig, Germany
}

\author[2]{Timo de Wolff\thanks{\email{t.de-wolfft@tu-braunschweig.de}}}

\author[2]{Dirk A. Lorenz\thanks{\email{d.lorenz@tu-braunschweig.de}}}

\maketitle

\begin{abstract}
  In this paper we consider the problem of learning variational models in the context of supervised learning via risk minimization. Our goal is to provide a deeper understanding of the two approaches of learning of variational models via bilevel optimization and via algorithm unrolling. The former considers the variational model as a lower level optimization problem below the risk minimization problem, while the latter replaces the lower level optimization problem by an algorithm that solves said problem approximately. Both approaches are used in practice, but unrolling is much simpler from a computational point of view. To analyze and compare the two approaches, we consider a simple toy model, and compute all risks and the respective estimators explicitly. We show that unrolling can be better than the bilevel optimization approach, but also that the performance of unrolling can depend significantly on further parameters, sometimes in unexpected ways: While the stepsize of the unrolled algorithm matters a lot (and learning the stepsize gives a significant improvement), the number of unrolled iterations plays a minor role.
\end{abstract}



\section{Introduction}
\label{sec:introduction}

In image and signal processing and beyond, quantities of interest are often reconstructed from potentially noisy observations by means of suitably parameterized energy minimization problems \cite{nikolova2011,kappes2013,kolmogorov2006,szeliski2006,boykov2001}. An unknown ground truth $\y$ is then approximately recovered in terms of an optimization problem
\begin{equation}\label{eq:general_energy_minimization_problem}
  \hat \y \in \argmin_{\z} f(\z, \xx, \w)
\end{equation}
where the energy $f$ to be minimized depends on an observation $\xx$ and parameters $\w$. Purely knowledge-driven approaches, for example based on sparsity assumptions \cite{daubechies2004,jin2012}, rely on the assumption that a suitable energy function and associated parameters are a priori known or can be hand-crafted from domain knowledge. Data aided approaches \cite{arridge2019,ongie2020,aggarwal2018} are commonly used in the opposite case where choosing an appropriate energy is not obvious and thus, the parameters $\w$ or parts of them shall be learned from data. In this work, we analyze the method of unrolling \cite{monga2021,chen2021b} for the learning of variational models \cite{scherzer2009,vese2016}. Here, the minimization problem (\ref{eq:general_energy_minimization_problem}) has the specific form
\begin{equation}\label{eq:variational_problem}
  \hat \y \in \argmin_{\z} \cS(\z,\xx) + \cR(\z,\w)
\end{equation}
with a similarity measure $\cS$ and a regularizer $\cR$. Moreover, we are given empirical data $(\xx_{i},\y_{i})$ for $i=1,\dots,m$ where each tuple includes an observation $\xx_{i}$ as well as the associated ground truth $\y_{i}$. Therefore, one aims to learn optimal parameters $\w$ by minimizing the empirical risk: For a loss function $\ell$, this amounts to solving the bilevel problem
\begin{equation}\label{eq:bilevel_problem}
\min_{\w}\ \tfrac1m\sum_{i=1}^{m}\ell(\hat \y_{i},\y_{i})\qquad \text{s.t.}\ \hat \y_{i} \in \argmin_{\z}\cS(\z,\xx_{i}) + \cR(\z,\w)\ .
\end{equation}
By learning parts of variational objectives, patterns in the data can be captured that elude human perception and therefore cannot find their way into hand designed and purely knowledge-driven objectives in classical approaches.

\subsection{Algorithm unrolling}
\label{sec:algorithm_unrolling}

Bilevel problems \cite{bracken1973,colson2007,sinha2017,bard2013,dempe2020} like \eqref{eq:bilevel_problem} are notoriously hard to solve. To apply gradient descent, the solution operator of the lower-level problem \eqref{eq:variational_problem} needs to be differentiated with respect to the parameters $\w$. Given that the objective of the lower-level problem is differentiable and attains a unique minimum, gradients can be computed by implicit differentiation \cite{ghadimi2018}. However, non-smooth objectives and such with potentially multi-valued solution operators are commonly used in image and signal-processing. The technique of \emph{unrolling} (see \cite{monga2021} for a recent survey and \cite{chen2021b} for a broader context) circumvents the need to differentiate the solution operator. Here, one considers an iterative algorithm which is able to solve the lower level problem and replaces the optimal solution $\hat \y_{i}$ of the lower level problem by the $N$-th iterate of this algorithm. If we denote this iterate by $A_{N}(\xx_{i},\w)$, learning by unrolling the lower level problem amounts to solving the surrogate problem
\begin{equation}\label{eq:unrolling_problem}
  \min_{w}\tfrac1m\sum_{i=1}^{m}\ell(A_{N}(\xx_{i},\w),\y_{i})\ .
\end{equation}
The same algorithmic scheme $A_{N}(\cdot, \w^{*})$ with learned parameters $\w^{*}$ is subsequently utilized to make predictions on unseen data.

\subsubsection{Comparison with neural networks}
\label{sec:comparison_with_nn}

Generally speaking, both (\ref{eq:bilevel_problem}) and (\ref{eq:unrolling_problem}) are supervised learning \cite{Hastie2009} problems since the actual goal is to learn a mapping from the space of observations to the space of ground truths based on a collection of input-output pairs. Artificial neural networks \cite{lecun2015,goodfellow2016} constitute a different and recently very popular class of parameterized models that are frequently used in the context of supervised learning problems. However, apart from all advantages that come along with neural networks and deep learning, some shortcomings still exist which may be more or less obstructive depending on the application at hand. First, neural network models and predictions are usually not interpretable without additional effort \cite{samek2019explainable}. In contrast, energy minimization problems are naturally interpretable as the objective that is minimized is often modeled from statistical principles in terms of priors and discrepancies. Second, the ability of neural networks to generalize to unseen data relies heavily on the availability of sufficient training data. Despite the potential use of regularization, training neural networks on small datasets is prone to overfitting \cite{dietterich1995}. Conversely, domain knowledge and priors can be incorporated in energy minimization problems by means of tailored objective function terms. As a consequence, energy minimization problems can generalize well also in small data regimes, while usually featuring considerably fewer parameters than artificial neural networks. Recent approaches combine energy minimization and deep learning by using modern neural network design patterns to learn regularizers for variational objectives \cite{kobler2020total}.

\subsubsection{Comparison with bilevel learning}
\label{sec:comparison_with_bilevel}

The unrolling of iterative algorithms can be motivated by the difficulty to solve some bilevel problems directly with gradient descent. Beyond that and in contrast to bilevel learning, unrolling offers a holistic view on training and inference as the same algorithmic scheme is utilized at both stages. Compared to classical model based and also bilevel approaches where iterative algorithms often require large numbers of iterations to succeed, unrolling can provide better convergence rates as model parameters are trained with regard to algorithmic efficiency.

\subsubsection{Application to variational models}
\label{sec:applications_to_variational_models}

A wide range of unrolling approaches with different applicational backgrounds have been derived based on variational models. In addition to the area of application that often motivates the concrete terms ${\cal R}$ and ${\cal S}$ used in (\ref{eq:variational_problem}), approaches can be distinguished in terms of further characteristics, for example, the unrolled algorithmic procedure $A$, associated algorithmic parameters like stepsizes and extrapolation factors, the number of unrolled iterations $N$, and finally, those parts of the objective which are parameterized and shall be learned. In the seminal work \cite{gregor2010}, the authors propose the learned iterative soft thresholding algorithm (LISTA) which is an unrolling of the ISTA algorithm from \cite{beck2009} and also Coordinate Descent \cite{li2009} to learn dictionaries for sparse coding \cite{olshausen1996}.  In \cite{brauer2019}, the Chambolle-Pock algorithm \cite{chambolle2011} is unrolled to learn an analysis operator and appropriate primal-dual stepsizes for speech dequantization, and it is shown that the resulting learned variational model outperforms the purely knowledge-driven variant introduced in \cite{brauer2016}. A different approach is pursued in \cite{adler2018} where the same algorithm is unrolled with trainable proximal operators in the context of tomographic reconstruction. In \cite{adler2017}, a similar approach is investigated that uses the Wasserstein loss \cite{cuturi2013} instead of the common $\ell_{2}$-loss. The authors of \cite{bertocchi2020} develop a network architecture for image restoration that is obtained by unrolling a proximal interior point algorithm \cite{bonettini2009} and learning associated barrier parameter, stepsize and penalization weight. The authors of \cite{ablin2019} investigate an architecture where only the stepsizes of ISTA are learned and demonstrate state-of-the art results given sparse solutions. Beyond the above-mentioned, there are numerous other recent works that consider or use the unrolling of variational models. As a complete list goes beyond the scope of this paper, we refer to the survey papers \cite{monga2021} and \cite{chen2021b}.

\subsubsection{Theoretical background}
\label{sec:theoretical_understanding}

If we focus exclusively on the attainable reconstruction error, fundamental questions arise that concern the expressivity of bilevel learning and unrolling as well as the suitability of both paradigms in view of specific applications and data distributions. In the case of unrolling, additional degrees of freedom, especially the utilized algorithm, algorithmic parameters, and the number of unrolled iterations, need to be taken into account.

It has been observed previously in \cite{brauer2019} that results can be rather insensitive with respect to the number of unrolled iterations. The paper \cite{takabe2020} investigates learned step-sizes in the case of gradient descent. It is shown that learned step-sizes can come close to the theoretically optimal step-sizes.
The paper \cite{chen2018} analyzes LISTA and provides insight about how the weights in different layers should be coupled to
guarantee convergence of the method. In a follow up \cite{liu2019a} the authors show that one can obtain a similar performance with analytic weights in contrast to learned weights for LISTA.
The paper~\cite{liu2019} analyzes an unrolled proximal method and shows that the proposed method can indeed be trained to obtain critical points of the true loss.
The paper~\cite{malezieux2022} applies unrolling to the problem of dictionary learning and particularly concludes that too many unrolled iterations lead to numerical instabilities while a few iterations lead to high performance. In particular, the authors observed that the calculation of gradients can become unstable and that a truncated backpropagation improves this behavior.

\subsection{Contribution}
\label{sec:contribution}

We investigate a toy model for learning variational models by bilevel optimization and unrolling of gradient descent. To do so, we first investigate the expressivity of the bilevel and the unrolling approach. This allows us to calculate best risks and the corresponding estimators for each approach exactly and thus a comparison of the two approaches. 

Even though the toy model is indeed quite simple, we can observe some (at least to us) surprising phenomena:
\begin{itemize}
\item In theory, the number of unrolled iterations plays a minor role if the stepsize is learned as well. However, the theoretical best risks differ substantially if an even or an odd number of iterations is unrolled. We are not aware that this phenomenon has been observed in practice and we suspect that steep minima in the odd case are responsible for this.
\item Optimal bilevel estimators may not exist in situations where optimal unrolling estimators exist.
\item The stepsize of the unrolled algorithm does influence the respective risk considerably. As a consequence, learning the stepsize as well is expected to be beneficial.
\end{itemize}
We will discuss these findings, their relations to other experiments and their relevance for practical problems in Section~\ref{sec:discussion}.

\subsection{Organization}
\label{sec:organization}

The remainder of this paper is structured as follows. In Section~\ref{sec:model} we introduce the framework for our theoretical investigations including a specific energy function and two data models. For the considered energy function we derive explicit formulas for the exact minimizer and for the $N$-th iterate of unrolled gradient descent. Based on that, we further derive specific risk functions to be minimized depending on both the utilized data model and whether exact bilevel optimization or unrolling shall be applied. In Section~\ref{sec:expressivity} we address the expressivity of exact bilevel optimization and unrolling, i.e., we derive sets of admissible estimators for both cases. Subsequently, we deduce best estimators and associated minimal risks in Section~\ref{sec:best_estimators}. A visualization and a further discussion of the previous results is given in Section~\ref{sec:discussion-visualization} and Section~\ref{sec:discussion}, respectively. In Section~\ref{sec:conclusion} we conclude the paper.

\subsection{Notation}
\label{sec:notation}

For a vector $\aaa\in\R^{n}$ we denote by $\diag(\aaa)$ the $n\times n$ diagonal matrix with $\aaa$ on the diagonal. For symmetric matrices $\SSS,\T$ we write $\T\mgeq \SSS$ if $\T-\SSS$ is positive semi-definite and $\T\mge \SSS$ if $\T-\SSS$ is positive definite. By $\norm{\xx}_{2}$ for some $\xx\in\R^{n}$ we denote the $2$-norm, and for a matrix $\T\in\R^{m\times n}$ we denote by $\norm{\T}_{F}$ the Frobenius norm which is induced by the standard matrix inner product, namely $\scp{\T}{\SSS} = \trace(\T^{\top}\SSS)$. Moreover, we denote by $\Eig(\T,\lambda) = \left\{\y\in\R^{n} \ \middle|\ \T\y = \lambda \y  \right\}$ the eigenspace of $\T$ for the eigenvalue $\lambda$.

\section{Model}
\label{sec:model}

We introduce our toy model and use the standard setup for supervised learning. We assume that paired data $(\xx,\y)$ consisting of objects $\xx$ and labels $\y$ is given following some distribution (which we describe below in Section~\ref{sec:data-model}). The goal is to predict $\y$ from $\xx$, i.e., we want to find a map $f$ which maps an object $\xx$ to its prediction $\hat \y := f(\xx)$ (see~\cite[Section 1.2]{CuZh2007}). Our learning model consists of the lower-level problem, the upper level problem and the model for the data, see~\eqref{eq:bilevel_problem}. In this paper we do not treat the problem of empirical risk minimization, but the minimization of the true risk, i.e., we do not minimize
\begin{align*}
  \tfrac1m\sum_{i=1}^{m}\ell(\hat \y_{i},\y_{i})
\end{align*}
but the true risk
\begin{align*}
  \E_{(\xx,\y)}\ell(\hat \y,\y),
\end{align*}
where the expectation is taken with respect to the distribution of the data $(\xx,\y)$.

\subsection{The data model}
\label{sec:data-model}

Our data model assumes that the given data $(\xx,\y)$ consists of noisy data $\xx$ which has been contaminated with additive noise, i.e., we assume that
\begin{align*}
  \xx = \y + \bm \epsilon
\end{align*}
with noise $\bm \epsilon\in\R^{n}$ and clean data $\y\in\R^{n}$. For simplicity we always assume that the components $\epsilon_{j}$ of noise are i.i.d. $\epsilon_{j}\sim\cN$ where $\cN$ is the normal distribution with mean zero and variance $\sigma^{2}$.

For the data $\y$ we assume two different but similar cases:
\begin{enumerate}
\item \textbf{Random constant vectors:} We assume that $\y = \lambda\1$ where $\lambda\in\R$ is distributed according to a distribution $\cD$ with expectation $\E(\lambda) = \mu$ and variance $\Var(\lambda) = \theta^{2}$ for some $\mu,\theta\in\R$.
\item \textbf{Random i.i.d. vectors:} We assume that all components $y_{j}$, $j=1,\dots,n$ are independent and identically distributed according to a distribution $\cD$ with expectation $\E(y_{j}) = \mu$ and variance $\Var(y_{j}) = \theta^{2}$ for some $\mu,\theta\in\R$.
\end{enumerate}
In the first case, we have one constant signal where the value of the signal is random, and in the second case we have a slightly noisy clear signal.

Our full data models have the following set of parameters:
\begin{quote}
  \begin{description}
  \item[$n$:] Ambient dimension.
  \item[$\sigma^2$:] Variance of the noise.
  \item[$\theta^2$:] Variance of the signal.
  \item[$\mu$:] Expected value of the signal. 
  \end{description}
\end{quote}

\subsection{Lower-level problem}
\label{sec:lowe-level}

Our lower level problem is  a very simple variational denoising model: For data $\xx\in\R^{n}$ we produce denoised data by solving
\begin{align}\label{eq:lower-level-prob}
  \hat \y = \argmin_{\z}\tfrac12\norm{\z-\xx}_{2}^{2} + \tfrac12\norm{\RRR\z}_{2}^{2}
\end{align}
where the parameters of the model are the coefficients of the matrix $\RRR\in\R^{k\times n}$. The first term $\tfrac12\norm{\z-\xx}_{2}^{2}$ is (up to a missing factor of $1/\sigma^{2}$) the negative log likelihood of the noise distribution. The second term $\tfrac12\norm{\RRR\z}_{2}^{2}$, however, is not the negative log-likelihood of the data distribution in both cases we consider. We use it here for convenience.

A calculation of the optimality condition of~\eqref{eq:lower-level-prob} shows that  the unique minimizer is given by
\begin{align}\label{eq:lower-level-prob-sol}
  \hat \y = (\I + \RRR^{\top}\RRR)^{-1}\xx.
\end{align}
As a minimization algorithm for this lower-level problem we consider simple gradient descent with constant stepsize $\omega>0$. If we initialize the gradient descent with $\z^{0}=\bm 0$ and define $\M\coloneqq \I - \omega (\I + \RRR^{\top}\RRR)$, then we get as $N$-th iterate
\begin{equation*}
  \begin{aligned}
    \z^{N} = \textstyle \M^{N}\z^{0} + \omega \sum_{j=0}^{N-1}\M^{j}\xx = \textstyle \omega \sum_{j=0}^{N-1}(\I - \omega (\I + \RRR^{\top}\RRR))^{j}\xx
  \end{aligned}
\end{equation*}
and further, using the geometric sum $\sum_{j=0}^{N-1}\A^{j} = (\I - \A)^{-1}(\I - \A^{N})$, we obtain
\begin{equation}
  \label{eq:lower-level-prob-unrolling}
  \z^{N} = (\I + \RRR^{\top}\RRR)^{-1}(\I - (\I - \omega(\I + \RRR^{\top}\RRR))^{N})\xx\ .
\end{equation}

\subsection{Upper level problem}
\label{sec:upper-level}
The objective in the upper level problem is simply the true risk for the squared loss
\begin{align*}
  \ell(\hat \y,\y) = \tfrac12\norm{\hat \y-\y}_2^2 \ .
\end{align*}
In principle, the loss should be motivated by the data model and other losses besides the squared loss are used (e.g. the cross entropy for regression problems) but we concentrate on the squared loss for simplicity. Note that the loss $\ell$ is in general not related to the similarity measure $\cS$ in the lower level problem.

Together with our data model we get the objective
\begin{align*}
  \E_{\substack{\bm \epsilon\sim\cN\\\y\sim\cD}}\tfrac12\norm{\hat \y - \y}_{2}^{2} \ .
\end{align*}

To calculate this risk, we note that both the solution operator of the lower level problem from~(\ref{eq:lower-level-prob-sol}) and the map from~(\ref{eq:lower-level-prob-unrolling}) (which gives the $N$-th iterate of gradient descent for the lower level problem) are linear operators. More precisely, we have 
\begin{align*}
  \T = (\I+\RRR^{\top}\RRR)^{-1}\in\R^{n\times n}
\end{align*}
for the solution operator and
\begin{align*}
  \T = (\I + \RRR^{\top}\RRR)^{-1}(\I-(\I-\omega(\I+\RRR^{\top}\RRR))^{N})\in\R^{n\times n}
\end{align*}
for unrolling. In both cases, the estimation is $\hat \y = \T\xx = \T(\y+\bm\epsilon)$ for some $\T\in\R^{n\times n}$ and 
hence, we aim to calculate
\begin{align*}
  \cE(\T) := \E_{\substack{\bm \epsilon\sim\cN\\\y\sim\cD}}\tfrac12\norm{\T(\y+\bm\epsilon) - \y}_{2}^{2}.
\end{align*}

\begin{lemma}[True risk for additive, independent noise]
  If $\y$ and $\bm\epsilon$ are independent and the noise has zero mean, it holds that
  \begin{align*}
    \cE(\T) = \E_{\y} \tfrac12\norm{(\T-\I)\y}_{2}^{2}  + \E_{\bm \epsilon}\tfrac12\norm{\T\bm\epsilon}_{2}^{2}.
  \end{align*}
\end{lemma}
\begin{proof}
  A calculation gives
\begin{align*}
  \E_{\substack{\bm \epsilon\sim\cN\\\y\sim\cD}} \tfrac12\norm{\T(\y+\bm\epsilon)-\y}_{2}^{2} & = \E_{\substack{\bm \epsilon\sim\cN\\\y\sim\cD}}\Big[\tfrac12\norm{(\T-\I)\y}_{2}^{2} + \scp{(\T-\I)\y}{\T\bm \epsilon} + \tfrac12\norm{\T\bm \epsilon}_{2}^{2}\Big]\\
                                                                                                    & = \E_{\y\sim\cD}\Big[\tfrac12\norm{(\T-\I)\y}_{2}^{2}\Big] + \scp{\E_{\y\sim\cD}(\T-\I)\y}{\E_{\bm \epsilon\sim\cN}\T\bm \epsilon} + \E_{\bm \epsilon\sim\cN}\tfrac12\norm{\T\bm \epsilon}_{2}^{2}\\
  & = \E_{\y\sim\cD}\Big[\tfrac12\norm{(\T-\I)\y}_{2}^{2}\Big]  + \E_{\bm \epsilon\sim\cN}\tfrac12\norm{\T\bm \epsilon}_{2}^{2},
\end{align*}
since $\E_{\bm \epsilon\sim\cN}\bm \epsilon = 0$.
\end{proof}
We see that the objective decouples into two terms: The noise term $\E_{\bm \epsilon}\tfrac12\norm{\T\bm \epsilon}_{2}^{2}$ and the data term $\E_{\y}\tfrac12\norm{(\T-\I)\y}_{2}^{2}$.

For the noise the following lemma shows that the expectation of $\norm{\T\bm \epsilon}_2^2$ can be expressed in terms of the Frobenius norm $\norm{\T}_{F}^{2}$ of $\T$.
\begin{lemma}\label{lem:expectation-noise-term}
  If the components of $\bm \epsilon$ are i.i.d. with mean zero and varaince $\sigma^{2}$ it holds that
  \begin{align*}
    \E_{\bm \epsilon\sim\cN}\norm{\T\bm \epsilon}_{2}^{2} = \sigma^{2}\norm{\T}_{F}^{2}.
  \end{align*}

\end{lemma}
\begin{proof}
  We denote the rows of $\T$ by $\ttt_{1},\dots,\ttt_{n}$ and the entries of $\T$ by $t_{ij}$ and get
  \begin{align*}
    \E_{\bm \epsilon\sim \cN}\norm{\T\bm \epsilon}_{2}^{2} & = \E_{\bm \epsilon\sim \cN} \sum_{i=1}^{n}\scp{\ttt_{i}}{\bm \epsilon}^{2} =  \sum_{i=1}^{n}\E_{\bm \epsilon\sim \cN}\scp{\ttt_{i}}{\bm \epsilon}^{2}
                                  = \sum_{i=1}^{n}\E_{\bm \epsilon\sim \cN}\left[\sum_{j=1}^{n}t_{ij}^{2}\bm \epsilon_{j}^{2} + 2\sum_{j\neq l}t_{ij}t_{il}\bm \epsilon_{j}\bm \epsilon_{l}\right]\\
                             & = \sum_{i=1}^{n}\sum_{j=1}^{n}t_{ij}^{2}\E_{\bm \epsilon\sim \cN}\bm \epsilon_{j}^{2} = \norm{\T}_{F}^{2}\sigma^{2}.
  \end{align*}
\end{proof}

For the data term we get different results for random constant vectors and i.i.d. vectors:
\begin{lemma}\label{lem:expectation-data-term}
  For the data term we obtain the following results:
  \begin{enumerate}
  \item \textbf{Random constant vectors:} For $\y = \lambda\1$ with $\E\lambda = \mu$ and
    $\Var(\lambda) = \theta^{2}$ it holds that
    \begin{align*}
      \E_{\y}\norm{(\T-\I)\y}_{2}^{2} = (\mu^{2}+ \theta^{2})\norm{(\T-\I)\1}_{2}^{2}.
    \end{align*}
  \item \textbf{Random i.i.d. vectors:} For $\y$ with $y_{j}$ i.i.d. with $\E(y_{j}) = \mu$ and $\Var(y_{j}) = \theta^{2}$ it holds that
    \begin{align*}
      \E_{\y}\norm{(\T-\I)\y}_{2}^{2} = \mu^{2}\norm{(\T-\I)\1}_{2}^{2} + \theta^{2}\norm{\T-\I}_{F}^{2}.
    \end{align*}
  \end{enumerate}
\end{lemma}
\begin{proof}
  Both parts are straightforward calculations.
  For the first part we use $\E(\lambda^{2}) = \Var(\lambda) + \E(\lambda)^{2} = \theta^{2}+\mu^{2}$ and obtain
  \begin{align*}
  	\E_{\y}\norm{(\T-\I)\y}_{2}^{2}= \E_{\lambda}\lambda^{2}\norm{(\T-\I)\1}_{2}^{2} = (\theta^{2}+\mu^{2})\norm{(\T-\I)\1}_{2}^{2}.
  \end{align*}

  \medskip 
  
  We show the second part.
  By independence of $y_{j}$ we have for symmetric $\M$ that
    \begin{align*}
      \E_{\y}\y^{\top}\M\y & = \E_{\y}\sum_{i,j}m_{ij}y_{i}y_{j} = \sum_{i\neq j}m_{ij}\E(y_{i})\E(y_{j}) + \sum_{i}m_{ii}\E(y_{j}^{2})\\
                    & = \mu^{2}\sum_{ij}m_{ij} + \theta^{2}\trace(\M).
    \end{align*}
    Hence, we get for $\M = (\T-\I)^{\top}(\T-\I)$
    \begin{align*}
      \E_{\y}\norm{(\T-\I)\y}_{2}^{2} & = \mu^{2}\sum_{ij}((\T-\I)^{\top}(\T-\I))_{ij} + \theta^{2}\trace((\T-\I)^{\top}(\T-\I))\\
                                  & = \mu^{2}\norm{(\T-\I)\1}_{2}^{2} + \theta^{2}\norm{\T-\I}_{F}^{2}.
    \end{align*}
\end{proof}

Combining the previous lemmata, we obtain the total risks for our two data models:
\begin{corollary}\label{cor:true-risks}
\mbox{}
  \begin{enumerate}
  \item \textbf{Random constant vectors:} For $\y = \lambda\1$ with $\E\lambda = \mu$ and
    $\Var(\lambda) = \theta^{2}$ it holds that
    \begin{align*}
      \cE_{\textup{const.}}(\T):= \cE(\T) = \tfrac{\mu^{2}+ \theta^{2}}2\norm{(\T-\I)\1}_{2}^{2} + \tfrac{\sigma^{2}}2\norm{\T}_{F}^{2}.
    \end{align*}
  \item \textbf{Random i.i.d. vectors:} For $\y$ with $y_{j}$ i.i.d. with $\E(y_{j}) = \mu$ and $\Var(y_{j}) = \theta^{2}$ it holds that
    \begin{align*}
      \cE_{\textup{i.i.d.}}(\T) := \cE(\T) = \tfrac{\mu^{2}}2\norm{(\T-\I)\1}_{2}^{2} + \tfrac{\theta^{2}}2\norm{\T-\I}_{F}^{2}+\tfrac{\sigma^{2}}2\norm{\T}_{F}^{2}.
    \end{align*}
  \end{enumerate}
\end{corollary}
\begin{proof}
  For (1) combine Lemma~\ref{lem:expectation-noise-term} and Lemma~\ref{lem:expectation-data-term} (1) and for (2) combine Lemma~\ref{lem:expectation-noise-term} with Lemma~\ref{lem:expectation-data-term} (2).
\end{proof}

\section{Expressivity}
\label{sec:expressivity}

In this section we analyze the expressivity of bilevel learning and unrolling for the lower level problem~(\ref{eq:lower-level-prob}). Since the resulting maps are linear in both cases, we just aim to characterize the set of linear maps that can be obtained by one of the two approaches.
We describe the set of bilevel estimators for the lower level problem (\ref{eq:lower-level-prob}) in Theorem~\ref{thm:expressivity-bilevel} and the set of unrolling estimators for the same lower level problem in Theorem~\ref{thm:expressivity-unrolling}.

\subsection{Expressivity of bilevel learning}
\label{sec:expressivity-of-bilevel}
As we have seen in Section~\ref{sec:model} in~(\ref{eq:lower-level-prob-sol}), bilevel estimators are of the form
\begin{align*}
  T = (\I + \RRR^{\top}\RRR)^{-1}
\end{align*}
with some $\RRR\in\R^{k\times n}$. Hence, the set of $n\times n$ matrices which can be expressed in this way depends on $k$ and we denote
\begin{align*}
  \cA_{k} := \{(\I+\RRR^{\top}\RRR)^{-1}\mid \RRR\in\R^{k\times n}\}.
\end{align*}
This set can be characterized with the help of the singular value decomposition of $\RRR$. The final result is:
\begin{theorem}\label{thm:expressivity-bilevel}
  The set of bilevel estimators for the lower level problem~(\ref{eq:lower-level-prob}) with matrices $\RRR\in \R^{k\times n}$ is
  \begin{align*}
    \cA_{k} = \{\T\in\R^{n\times n}\mid \dim(\Eig(\T,1))\geq n-k,\ \T^{\top}=\T,\ \bm 0\mle \T\mleq \I\}.
  \end{align*}
\end{theorem}
\begin{proof}
  Let $\T = (\I + \RRR^{\top}\RRR)^{-1}\in {\cal A}_{k}$. The matrix $\I + \RRR^{\top}\RRR$ is symmetric and positive definite and hence, the same holds for $\T$. As $\RRR^{\top}\RRR$ is positive semidefinite, there exist an orthonormal matrix $\V\in\R^{n\times n}$ and a diagonal matrix $\Sigmaa = \diag(\sigma_{1},\ldots,\sigma_{n})\in \R^{n\times n}$ with $\sigma_{1}\geq \dots \geq \sigma_{k} \geq 0$ and $\sigma_{k+1} = \dots = \sigma_{n} = 0$ such that $\RRR^{\top}\RRR = \V\Sigmaa\V^{\top}$. It follows that
  \begin{align*}
    \T & = (\I + \RRR^{\top}\RRR)^{-1} = (\I + \V\diag(\sigma_{1},\ldots,\sigma_{n})\V^{\top})^{-1} \\
    & = (\V(\I + \diag(\sigma_{1},\ldots,\sigma_{n})))\V^{\top})^{-1} = \V\diag\left(\tfrac{1}{1 + \sigmaa}\right)\V^{\top}.
  \end{align*}
  Therefore, we have $\T\in \{\T\in\R^{n\times n}\mid \dim(\Eig(\T,1))\geq n-k,\ \T^{\top}=\T,\ 0\mle \T\mleq \I\}$.
  \medskip

  Vice versa, if $\T$ fulfills $\dim(\Eig(\T,1))\geq n-k$, $\T^{\top}=\T$ and $\bm 0\mle \T\mleq \I$, then we can write $\T = \V\Thetaa\V^{\top}$ with $\Thetaa = \diag(\thetaa)$ and $0<\theta_{j}\leq 1$. We define $\Sigmaa = \diag(\sigma_{1},\ldots,\sigma_{n})$ by setting $\sigma_{j} = \tfrac{1}{\theta_{j}}-1$ and get that $\T = (\I + \V\Sigmaa \V^{\top})^{-1}$. Hence, any $\RRR = \U\Sigmaa \V^{\top}$ with some orthonormal $\U$ gives $\T = (\I+\RRR^{\top}\RRR)^{-1}$ as desired.
\end{proof}

\subsection{Expressivity of unrolling}
\label{sec:expressivity-unrolling}

In the unrolling case, we have seen in~(\ref{eq:lower-level-prob-unrolling}) that the estimators are of the form
\begin{align*}
  \U = (\I+\RRR^{\top}\RRR)^{-1}(\I-(\I-\omega(\I+\RRR^{\top}\RRR))^{N})
\end{align*}
and hence, the expressivity depends on the depth $N$ that we unroll and also on the stepsize $\omega$ of the algorithm. The set of possible estimators is
\begin{align*}
  \cB_{N,k,\omega} := \left\{(\I+\RRR^{\top}\RRR)^{-1}(\I-(\I-\omega(\I+\RRR^{\top}\RRR))^{N}) \mid \RRR\in\R^{k\times n}\right\}.
\end{align*}
An analysis similar to the previous section leads to the following result:
\begin{theorem}\label{thm:expressivity-unrolling}
  The set of unrolling estimators for the lower level problem~(\ref{eq:lower-level-prob}) unrolled for $N$ steps of gradient descent with stepsize $\omega>0$ and matrices $\RRR\in\R^{k\times n}$ is as follows:
  \begin{enumerate}
  \item $N$ even:
    \begin{align*}
      \cB_{N,k,\omega} & =\left\{\U\in\R^{n\times n}\ \middle|\ \parbox{7.5cm}{$\U=\U^{\top},\ \dim(\Eig(\U,1-(1-\omega)^{N})))\geq n-k,\\ \U \mleq [1-(1-\omega)^{N}]\I$}\right\}
    \end{align*}
  \item $N$ odd: There exists a constant $c_{N,\omega}$ which fulfills 
    \begin{align*}
      \omega\Big(\tfrac12 + \tfrac1{N+1}\Big) \leq c_{N,\omega} &\leq \omega\Big(\tfrac12 + \tfrac1N\Big(\tfrac{1+\log(N)/2}{2-\tfrac{\log(N)}N}\Big)\Big) && \text{if}\  ((N-1)\omega+1)(1-\omega)^{N-1}\geq 1,\\
      c_{N,\omega} & = 1-(1-\omega)^{N}&& \text{else,}
    \end{align*}
    such that
    \begin{align*}
      \cB_{N,k,\omega} =
      \{\U\in\R^{n\times n}\mid \U=\U^{\top},\ \dim(\Eig(\U,1-(1-\omega)^{N}))\geq n-k,\ c_{N,\omega}\I\mleq \U\}.
    \end{align*}
  \end{enumerate}
\end{theorem}
\begin{proof}
  As in the proof of Theorem~\ref{thm:expressivity-bilevel} we write $\RRR^{\top}\RRR = \V^T\Sigmaa \V$ with $\Sigmaa = \diag(\sigma_{1},\ldots,\sigma_{n})$. We denote $\sigmaa = (\sigma_{1},\dots,\sigma_{n})$ and abbreviate $\diag(\sigmaa)$ for vectors $\sigmaa$. Any $\U\in \cB_{N,k,\omega}$ can be written as
  \begin{align*}
    \U = \V^{\top}\diag(f_{N,\omega}(\sigmaa))\V\quad\text{with}\quad 
    f_{N,\omega}(\sigma) = \tfrac{1-(1-\omega(1+\sigma))^{N}}{1+\sigma},
  \end{align*}
  where we applied $f$ componentwise to the vector $\sigmaa$ with entries ordered decreasingly. More precisely we have  (since the rank of $\RRR$ is $k$) $\sigma_{1}\geq \sigma_{2}\geq\cdots \geq\sigma_{k}\geq \sigma_{k+1}= \cdots=\sigma_{n}=0$ .
  In other words, $\U$ is a symmetric matrix which has $f_{N,\omega}(0)$ as eigenvalue with multiplicity at least $n-k$ and $k$ further eigenvalues that are in the range of $f_{N,\omega}:\aoival{0}{\infty}\to \R$.

  Thus, we only need to analyze the range of $f_{N,\omega}$: If $N$ is
  even and $\omega>0$ is arbitrary, the function $f_{N,\omega}$ is
  strictly decreasing and unbounded from below with
  $f_{N,\omega}(0) = 1-(1-\omega)^{N}$. Hence, it holds that
  $f_{N,\omega}:\aoival{0}{\infty}\to
  \oaival{-\infty}{1-(1-\omega)^{N}}$ is onto.

  In the case of odd $N$ the situation is more complicated: We still
  have $f_{N,\omega}(0) = 1-(1-\omega)^{N}$, but if $f_{N,\omega}'(0)\leq 0$, then 
  $f_{n,\omega}:\aoival{0}{\infty}\to\R$ has a single local minimum
  which is positive, but the function is unbounded from above. Hence we
  have that there is some positive value $c_{N,\omega}$ such that
  $f_{N,\omega}:\aoival{0}{\infty}\to\aoival{c_{N,\omega}}{\infty}$ is onto.
  If $f_{N,\omega}'(0)$ is positive, the function is strictly increasing and its minimum is at $f_{N,\omega}(0)$. Computing $f_{N,\omega}'(0)$ shows that this is the case exactly when $\big((N-1)\omega+1\big)(1-\omega)^{N-1}< 1$.

  Now we derive upper and lower bounds for $c_{N,\omega}$ in the case $((N-1)\omega+1)(1-\omega)^{N-1}\geq 1$. We substitute $r = 1 - \omega(1+\sigma)$ (remember that $\sigma>0$, i.e., $r\leq 1-\omega$). Then we have that $f_{N,\omega}(\sigma)  = \omega\tfrac{1-r^{N}}{1-r}$, i.e., we only need to estimate the minimum of $h_{N}(r) = (r^{N}-1)/(r-1)$ from above and from below.
  
  First we note that any function value of $h_{N}$ is an upper bound for the minimum, i.e.,  
  we can estimate the minimum from above by evaluating
  $(1-r^{N})/(1-r)$ at $r^{*} = -1 + \log(N)/N$, which gives the
  upper bound
  \begin{align*}
    \frac{(-1+\tfrac{\log(N)}{N})^{N}-1}{\tfrac{\log(N)}{N}-2} = \frac{1+(1-\tfrac{\log(N)}{N})^{N}}{2-\tfrac{\log(N)}{N}}.
  \end{align*}
  We further estimate
  \begin{align*}
    \left(1-\tfrac{\log(N)}{N}\right)^{N} = \exp\left(N\log(1-\tfrac{\log(N)}{N})\right) \leq \tfrac1N,
  \end{align*}
  where we used $\log(1-x)\leq -x$. Hence, we get the claimed upper bound
  \begin{align*}
    b_{N} = \frac{1+\tfrac1N}{2-\tfrac{\log(N)}{N}} = \frac12 + \frac1N\Big(\frac{1+\log(N)/2}{2-\tfrac{\log(N)}N}\Big).
  \end{align*}
  
  Now we show that $a_{N}= \tfrac12 + \tfrac1{N+1}$ is a lower bound for $h_{N}$. We prove this by induction: First we rewrite $h_{N}(r) = r^{N-1}+\cdots + r + 1$. A lower bound $a_{N}$ is any value such that $h_{N}(r)-a_{N}$ is a non-negative polynomial. For $N=3$ we have $h_{3}(r)-a_{3} = r^{2}+r+1-a_{3}$. Recalling the fact that
  \begin{align}\label{eq:pos-square}\tag{*}
    ar^{2}+r+c\geq 0 \iff a\geq 0 \text{ and } ac\geq \tfrac14,
  \end{align}
  we see that $1-a_{3}= \tfrac14$, i.e., $a_3 = \tfrac34$ is the optimal lower bound.
  
  Now assume that $h_{N}\geq a_{N}$ is proven. We have
  \begin{align*}
    h_{N+2}(r) &= r^{N+1} + r^{N} + \cdots +r+1 = r^{2}h_{N}(r)  + r +1 \geq a_{N}r^{2}+r+1. 
  \end{align*}
  Using~(\ref{eq:pos-square}), we see that $a_{N}r^{2}+r+1-a_{N+2}$ is non-negative if
  \begin{align*}
    a_{N}(1-a_{N+2})= \tfrac14,
  \end{align*}
  hence $a_{N+2} = 1 - \tfrac1{4a_{N}} = \tfrac12 + \tfrac{1}{N+3}$ as claimed.
\end{proof}

This theorem has important consequences:
\begin{itemize}
\item The expressivity of unrolling changes greatly if the number of unrolled iterations is even or odd: In the odd case, only positive definite estimators are possible but their norm may be unbounded, while in the even case, all estimators have a norm equal to $1-(1-\omega)^{N}$ but may be singular and/or have negative eigenvalues.
\item In the even case, increasing the number of unrolled iterations (with fixed $k$) does not increase the expressivity if the stepsize $\omega$ is adjusted properly, in other words: unrolling $2$ iterations is as good as unrolling $4$ iterations. We will see this more clearly in Section~\ref{sec:best-unrolling}.
\end{itemize}

We note furthermore that the proof of nonnegativity of $h_N(r) - a_N$ shown above can be interpreted from a more abstract viewpoint: Deciding nonnegativity of polynomials is a classical key problem in real algebraic geometry with strong ties to polynomial optimization; see e.g., \cite{Blekherman:Parrilo:Thomas,Lasserre:IntroductionPolynomialandSemiAlgebraicOptimization}.
Nonnegativity is frequently shown by what is called a certificate of nonnegativity.
One of these certificates are circuit polynomials introduced in \cite{Iliman:deWolff:Circuits}, and what is effectively done in the proof above is a decomposition of $h_N(r) - a_N$ into a sum of nonnegative circuit polynomials; see also e.g. \cite{Iliman:deWolff:GP}.

\section{Best risks and estimators}
\label{sec:best_estimators}
In this section,  after calculating the best linear estimators, we aim for the best unrolling and bilevel estimators for both data models.  For this purpose we minimize the true risks from Corollary~\ref{cor:true-risks} over linear operators of the corresponding form.

We provide an infimum for the risks of the best bilevel estimator in the case of random constant vectors, Theorem~\ref{thm:best-bilevel-constant}, and compute the minimum for this risk in the case of random i.i.d. vectors, Theorem~\ref{thm:best-bilevel-iid}. Also, we compute the minimum for the risks of the unrolling estimator in the case of random constant vectors, Theorem~\ref{thm:best-unrolling-constant}, and in the case of random i.i.d. vectors, Theorem~\ref{thm:best-unrolling-iid}. In our analysis we do consider all parameters varying. However, note that due to the expressivity results Theorem~\ref{thm:expressivity-bilevel} and Theorem~\ref{thm:expressivity-unrolling} we do not consider the case $k>n$ (i.e., $R$ having more rows than columns) as this case has exactly the same expressivity as the case $k=n$.

\subsection{Best linear estimators and their risks}
\label{sec:best-linear}
For now we focus on general linear operators, i.e., $\T \in \R^{n \times n}$. 
\subsubsection{Random constant vectors}
\label{sec:linear-const}
In the situation of random constant vectors, in which the true risk is given by 
\begin{align*}
  \cE_{\textup{const.}}(\T) = \tfrac{\mu^{2}+ \theta^{2}}2\norm{(\T-\I)\1}_{2}^{2} + \tfrac{\sigma^{2}}2\norm{\T}_{F}^{2},
\end{align*}
we get the following result: 
\begin{lemma}\label{lem:best-est-constant}
  The best linear estimator is $\T^{*}=\tfrac{\mu^{2}+\theta^{2}}{n(\mu^{2}+\theta^{2}) + \sigma^{2}}\1\1^{\top}$ and its corresponding risk is 
  \begin{align*}
    \min_{\T}{\cE_{\textup{const.}}(\T)} = \cE_{\textup{const.}}(\T^{*})= \tfrac{\sigma^{2}}{2}\tfrac{n(\mu^{2}+\theta^{2})}{n(\mu^{2}+\theta^{2})+\sigma^{2}}.
  \end{align*}	 
  In particular, the estimate for the denoised data of a given $\xx$ is 
  \begin{align*}
    \hat{\y}=\T^{*}\xx=\tfrac{\scp{\xx}{\1}(\mu^{2}+\theta^{2})}{n(\mu^{2}+\theta^{2}) + \sigma^{2}}\1.
  \end{align*}
\end{lemma}
The proof is given in \ref{app:further-proofs}.

Since in this data model the vectors are just constant vectors, one might naively expect that the optimal linear estimator will just be the estimator $T^{\text{ave}} = \tfrac1n \1\1^{T}$ which averaged all entries. The theorem shows that a slightly damped average (depending on the parameters of data and noise) will lead to a smaller risk and this shows that learned algorithms (which do not need to know the parameters of the data distribution) can outperform naive ad hoc estimators.

\subsubsection{Random i.i.d. vectors}
\label{sec:linear-iid}
Recall that Corollary~\ref{cor:true-risks} gives the true risk in the situation of random i.i.d. vectors as
\begin{align*}
  \cE_{\textup{i.i.d.}}(\T)=\tfrac{\mu^{2}}2\norm{(\T-\I)\1}^{2} + \tfrac{\theta^{2}}2\norm{\T-\I}_{F}^{2} +  \tfrac{\sigma^{2}}2\norm{\T}_{F}^{2}.
\end{align*}
\begin{lemma}\label{lem:best-est-iid}
  The best linear estimator in the case of i.i.d. vectors, i.e., the minimizer of $\cE_{\textup{i.i.d.}}$, is $\T^{*} = \tfrac{\theta^{2}}{\theta^{2}+\sigma^{2}} \I +  \tfrac{\sigma^{2}}{\theta^{2}+\sigma^{2}}\tfrac{\mu^{2}}{n\mu^{2} + \theta^{2}+\sigma^{2}}\1\1^{\top}$ and its corresponding risk is 
  \begin{align*}
    \min_{\T}{\cE_{\textup{i.i.d.}}(\T)} = \cE_{\textup{i.i.d.}}(\T^{*})&=\tfrac{\sigma^{2}}2\Big[\tfrac{n\mu^{2}\sigma^{2}}{(n\mu^{2} + \theta^{2} + \sigma^{2})^{2}}\Big(1 + \tfrac{n\mu^{2}\sigma^{2}}{(\theta^{2} + \sigma^{2})^{2}}\Big)\\ 
  & \quad + \tfrac{\theta^{2}}{(\theta^{2} + \sigma^{2})^{2}}\Big(\tfrac{n\sigma^{2}\mu^{2}}{n\mu^{2} + \theta^{2} + \sigma^{2}} + n\theta^{2} + \sigma^{2}\Big((n-1) + \tfrac{\theta^{2} + \sigma^{2}}{(n\mu^{2} + \theta^{2} + \sigma^{2})^{2}}\Big)\Big)\Big].
  \end{align*}	 
\end{lemma}
The proof is given in \ref{app:further-proofs}.

In this case the optimal estimator is not a damped weighted average, but a convex combination of a damped average and the identity.

\subsection{Best bilevel estimators and their risks}
\label{sec:best-bilevel}
Now we consider the bilevel approach, i.e., we aim to minimize the risk $\cE(\T)$ over maps of the form \begin{align*}
\T=(\I+\RRR^{\top}\RRR)^{-1}
\end{align*} 
with $\RRR \in \R^{k\times n}$. We begin with the situation of random constant vectors, but before we do so, we state a lemma that will help later:
\begin{lemma}\label{lem:unrolling_orthonormal}
  Let $\aaa\in \R^{n}$, $c_{j}\geq 0$ for $j=1,\dots,n$ and consider matrices $\V = [\vv_{1},\dots,\vv_{n}]$. Then there exists a solution of 
  \begin{align*}
    \min_{\V}\left\{ \sum_{j=1}^{n} \abs{\scp{\vv_{j}}{\aaa}}^{2}c_{j} \ : \ \sum_{j=1}^{n} \abs{\scp{\vv_{j}}{\aaa}}^{2} = \norm{\aaa}^{2},\ \V \text{ orthonormal} \right\}
  \end{align*}
  with $\vv_{j^{*}} = \tfrac{\aaa}{\norm{\aaa}}$ and $\vv_{j}\perp \aaa$ for $j \neq j^{*}$ with $j^{*}\in \argmin_{j}c_{j}$.
\end{lemma}
\begin{proof}
  We reformulate the minimization problem as a linear program. To that end we substitute $s_{j} = \abs{\scp{\vv_{j}}{\aaa}}^{2}$ and consider
   \begin{equation*}
     \begin{array}{ll@{}ll}
       \text{minimize}_{\sss} &\cc^{\top}&\sss &\\
       \text{subject to} & \sum\limits_{j}&{s_{j}} = \norm{\aaa}^{2} &\\
                                    & 0\leq &s_{j} \leq \norm{\aaa}^{2} & \text{ for all } j = 1,\ldots, n
     \end{array}
   \end{equation*}
   with $\cc = (c_1,\ldots,c_n)$ and $c_{j}\geq 0$. 
   An optimal solution of this linear program is given by a vertex of the feasible set. Thus, it is $\sss^{*} = \norm{\aaa}^{2}\cdot \e_{j^{*}}$ for some $j^{*} \in \{1,\ldots,n\}$ where $\e_{k}$ denotes the $k$-th vector of the $n$-dimensional Euclidean standard basis. It remains to show that there is an orthonormal basis $(\vv_{1},\dots,\vv_{n})$ with $\abs{\scp{\vv_{j}}{\aaa}}^{2} = s^{*}_{j}$. This is equivalent to  $\abs{\scp{\vv_{j^{*}}}{\aaa}}^{2} = \norm{\aaa}^{2}$ and $\abs{\scp{\vv_{j}}{\aaa}}^{2} = 0$ for $j\neq j^{*}$ and holds for $\vv_{j^{*}} = \tfrac{\aaa}{\norm{\aaa}}$ and $\vv_{j}\perp \aaa$ for $j \neq j^{*}$. Since $c_{j}\geq 0$ for all $j$, it is $j^{*}\in \argmin_{j}c_{j}$.
\end{proof}

\subsubsection{Random constant vectors}
\label{sec:bilevel-const}
In the situation of random constant vectors we want to minimize $\cE_{\textup{const.}}(\T)$ over maps $\T \in \cA_{k}$, i.e., we want to minimize 
\begin{align*}
  \tfrac{\mu^{2}+ \theta^{2}}2\norm{((\I+\RRR^{\top}\RRR)^{-1}-\I)\1}_{2}^{2} + \tfrac{\sigma^{2}}2\norm{(\I+\RRR^{\top}\RRR)^{-1}}_{F}^{2} \ .
\end{align*}

\begin{theorem}\label{thm:best-bilevel-constant}
  It holds that 
  \begin{align*}
    \inf_{\T\in \cA_{k}}{\cE_{\textup{const.}}(\T)} = \begin{cases}\tfrac{\sigma^{2}}{2}(n-k), & \text{if $k<n$}, \\
    \tfrac{\sigma^{2}}{2}\tfrac{n(\mu^{2}+\theta^{2})}{n(\mu^{2}+\theta^{2})+\sigma^{2}}, & \text{if $k=n$} \end{cases}
  \end{align*}
  but the infimum is not attained in both cases. Hence,  the best bilevel estimator does not exist and for the risk of any bilevel estimator $\hat{\T}$ it holds that 
  \begin{align*}
    \cE_{\textup{const.}}(\hat{\T})> 
    \begin{cases}
      \tfrac{\sigma^{2}}{2}(n-k), & \text{if $k<n$}, \\
      \tfrac{\sigma^{2}}{2}\tfrac{n(\mu^{2}+\theta^{2})}{n(\mu^{2}+\theta^{2})+\sigma^{2}}, & \text{if $k=n$.}
    \end{cases}
  \end{align*}
\end{theorem}
The proof is given in \ref{app:further-proofs}.
Interpreting the result above, a best bilevel estimator for random constant vectors does not exist. However, there exist estimators which give a risk arbitrarily close to $\sigma^{2}(n-k)/2$.

\subsubsection{Random i.i.d. vectors}
\label{sec:bilevel-iid}
In the situation of random i.i.d. vectors we want to minimize $\cE_{\textup{i.i.d.}}(\T)$ over maps $\T \in \cA_{k}$, i.e., we want to minimize 
\begin{align*}
\tfrac{\mu^{2}}{2}\norm{((\I+\RRR^{\top}\RRR)^{-1}-\I)\1}_{2}^{2} + \tfrac{\theta^{2}}2\norm{(\I+\RRR^{\top}\RRR)^{-1}-\I}_{F}^{2}+\tfrac{\sigma^{2}}2\norm{(\I+\RRR^{\top}\RRR)^{-1}}_{F}^{2}.
\end{align*}
In this case, best estimators do exist, but for the sake of readability we just state the best risks in the next theorem:
\begin{theorem}\label{thm:best-bilevel-iid}
  It holds that the attained minimal risk is given by 
  \begin{align*}
   \min_{\T\in\cA_{k}} \cE_{\textup{i.i.d.}}(\T) = 
   \begin{cases}
    \tfrac{\sigma^{2}}{2}\Big(k\tfrac{\theta^{2}}{\theta^{2}+\sigma^{2}} + (n-k)\Big), \quad \text{if } k<n \\
    \tfrac{\sigma^{2}}2\Big((n-1)\tfrac{\theta^{2}}{\theta^{2} + \sigma^{2}} + \tfrac{n\mu^{2}+\theta^{2}}{n\mu^{2} + \theta^{2} + \sigma^{2}}\Big), \quad \text{if } k = n \ .
    \end{cases}
  \end{align*}
\end{theorem}

We do not prove this theorem here, since we get the result as a byproduct in the proof of the next proposition.
\begin{proposition}\label{prop:best-bilevel-iid_estimators}
  The optimal risk in Theorem~\ref{thm:best-bilevel-iid} for $k<n$ is attained for
    $\RRR = \U\Lambdaa \V^{\top}$
  with $\U\in\R^{k\times n}$ with orthonormal rows ($\U\U^{\top}=\I_{k}$), $\Lambdaa = \tfrac{\sigma}{\theta}
  \begin{bmatrix}
    \I_{k} & \bm 0\\
    \bm 0 & \bm 0_{n-k}
  \end{bmatrix}$ and orthonormal $\V =
  \begin{bmatrix}
    \vv_{1} & \cdots & \vv_{n}
  \end{bmatrix}$ with $\vv_{1},\dots,\vv_{k}\bot \1$. The respective linear estimator is
  \begin{align*}
    \T^{*} = (\I+\RRR^{\top}\RRR)^{-1} = \V\Sigmaa \V^{\top},\ \Sigmaa = \diag(\tfrac{\theta^{2}}{\theta^{2} + \sigma^{2}},\dots,\tfrac{\theta^{2}}{\theta^{2} + \sigma^{2}},1,\dots,1) \ .
  \end{align*}
The optimal risk in Theorem~\ref{thm:best-bilevel-iid} for $k=n$ is attained for
  $\RRR = \U\Lambda \V^{\top}$
  with $\U\in\R^{n\times n}$ with orthonormal rows, $\Lambdaa = 
  \begin{bmatrix}
     \tfrac{\sigma}{\theta}\I_{n-1}& \bm 0\\
    \bm 0 & \tfrac{\sigma}{\sqrt{n\mu^{2} + \theta^{2}}}
  \end{bmatrix}$ and orthonormal $\V =
  \begin{bmatrix}
    \vv_{1} & \cdots & \vv_{n}
  \end{bmatrix}$ with $\vv_{2},\dots,\vv_{n}\bot \1$. The respective linear estimator is
  \begin{align*}
    \T^{*} = (\I+\RRR^{\top}\RRR)^{-1} = \V\Sigmaa \V^{\top},\ \Sigmaa = \diag(\tfrac{\theta^{2}}{\theta^{2} + \sigma^{2}}, \dots,\tfrac{\theta^{2}}{\theta^{2} + \sigma^{2}}, \tfrac{n\mu^{2} + \theta^{2}}{n\mu^{2} + \theta^{2} + \sigma^{2}})\ .
  \end{align*}
\end{proposition}
The proof is given in \ref{app:further-proofs}.

Thus, we have proven Proposition~\ref{prop:best-bilevel-iid_estimators} as well as Theorem~\ref{thm:best-bilevel-iid}. Note that the case of i.i.d. vectors reduces to the case of random constant vectors for $\theta\to 0$ (i.e., the variance of the entries vanishes, rendering the vector constant): The respective minimal risk from Theorem~\ref{thm:best-bilevel-iid} converges to the one on Theorem~\ref{thm:best-bilevel-constant} and the optimal $\RRR$ diverges for $\theta\to 0$.

\subsection{Best unrolling estimators and their risks}
\label{sec:best-unrolling}
For the rest of the section we consider unrolling estimators. In this case we aim to find the minimizer of the risk over maps of the form
\begin{align}\label{eq:estimator-unrolling}
  \T = \omega\sum_{j=0}^{N-1}(\I-\omega(\I+\RRR^{\top}\RRR))^{j} = (\I+\RRR^{\top}\RRR)^{-1}(\I-(\I-\omega(\I+\RRR^{\top}\RRR))^{N}).
\end{align}
with $\RRR \in \R^{k \times n}$.

Using functional calculus for operators, we get that if $\RRR^{\top}\RRR = \V\Sigmaa \V^{\top}$ with orthormal $\V$ and $\Sigmaa = \diag(\sigma_{1},\ldots,\sigma_{n})$ (where $\sigma_{k+1},\dots,\sigma_{n}=0$ if $\RRR$ is $k\times n$), then $\T$ from~(\ref{eq:estimator-unrolling}) is of the form
\begin{align} \label{eq:unrolling_estimator_svd}
  \T = \V f(\Sigmaa)\V^{\top},\ \text{with}\ f(\Sigmaa) := \diag(f(\sigma_{1}),\dots,f(\sigma_{n})),\ \text{and}\ f(s) =  \tfrac{1 - (1-\omega(1 + s))^{N}}{1+s}.
\end{align}

For convenience we define the following abbreviations of terms that appeared in previous expressions and will appear in the next results as well:
\begin{align*}
  \rho_{N,\omega} & := 1-(1-\omega)^{N}, & C_{1} & := \tfrac{(\mu^{2} + \theta^{2})n}{(\mu^{2} + \theta^{2})n + \sigma^{2}} & C_{2} & := \tfrac{\mu^{2}n + \theta^{2}}{\mu^{2}n + \theta^{2} + \sigma^{2}}\\
  C_{3} & := \tfrac{\theta^{2}}{\theta^{2} + \sigma^{2}}, & C_{i, \min} & := \min\{C_{i},\rho_{N,\omega}\}, & C_{i, \max} & :=\max\{C_{i},c_{N,\omega}\}\ , i\in \{1,2,3\}
\end{align*}
with $c_{N,\omega}$ from Theorem~\ref{thm:expressivity-unrolling}. 

\subsubsection{Random constant vectors}
\label{sec:unrolling-const}
\begin{theorem}\label{thm:best-unrolling-constant}
  Let $N\in\N$ and $0<\omega<2$ be fixed. For $N$ even and $k<n$ it holds that
  \begin{align*}
    \min_{\T\in\cB_{N,k,\omega}} \cE_{\textup{const.}}(\T) = \tfrac{\mu^{2} + \theta^{2}}{2}n(\rho_{N,\omega} - 1)^{2} + \tfrac{\sigma^{2}}{2}(n-k)\rho_{N,\omega}^{2} \ .
  \end{align*}
    For $k=n$ we have 
  \begin{align*}
    \min_{\T \in \cB_{N,k,\omega}}\cE_{\textup{const.}}(\T) &= \tfrac{\mu^{2} + \theta^{2}}{2}n(C_{1, \min}-1)^{2} + \tfrac{\sigma^{2}}{2}C_{1, \min}^{2}\ .
  \end{align*}
  The minimal risk is attained in both cases. 
  For $N$ odd and $k<n$ it holds for the minimal risk 
  \begin{align*}
   \min_{\T\in\cB_{N,k,\omega}} \cE_{\textup{const.}}(\T) &= \tfrac{\sigma^{2}}{2}((k-1)c_{N,\omega}^{2} + (n-k)\rho_{N,\omega}^{2})\\ 
   &\quad+ \min\Big\{\tfrac{\mu^{2} + \theta^{2}}{2}n(C_{1, \max} - 1)^{2} + \tfrac{\sigma^{2}}{2}C_{1, \max}^{2},\tfrac{\mu^{2} + \theta^{2}}{2}n(\rho_{N,\omega}-1)^{2} + \tfrac{\sigma^{2}}{2}c_{N,\omega}^{2}\Big\} \ .
  \end{align*}
  For $k=n$ we have 
    \begin{align*}
    \min_{\T \in \cB_{N,k,\omega}}\cE_{\textup{const.}}(\T) &= \tfrac{\mu^{2} + \theta^{2}}{2}n(C_{1, \max}-1)^{2} + \tfrac{\sigma^{2}}{2}C_{1, \max}^{2} + \tfrac{\sigma^{2}}{2}(n-1)c_{N,\omega}^{2}.
  \end{align*}
  Again, in both cases the minimal risk is attained. 
\end{theorem}
The proof is given in \ref{app:further-proofs}.

\subsubsection{Random i.i.d. vectors}
\label{sec:unrolling-iid}

For the case of random i.i.d. vectors we have the following result:

\begin{theorem}\label{thm:best-unrolling-iid}
Let $N\in\N$ and $0<\omega<2$ be fixed. For $N$ even and $k<n$ it holds that
  \begin{align*}
    \min_{\T\in\cB_{N,k,\omega}} \cE_{\textup{i.i.d.}}(\T) &= \tfrac{\mu^{2}}{2}n(\rho_{N,\omega} -1)^{2} + \tfrac{\theta^{2}}{2}(k-1)(C_{3,\min} - 1)^{2} + \tfrac{\theta^{2}}{2}(n-k)(\rho_{N,\omega}-1)^{2}\\
    &\quad + \tfrac{\sigma^{2}}{2}(k-1)C_{3,\min}^{2} + \tfrac{\sigma^{2}}{2}(n-k)\rho_{N,\omega}^{2}\\ 
    &\quad + \min\Big\{\tfrac{\theta^{2}}{2}(\rho_{N,\omega} -1)^{2} + \tfrac{\sigma^{2}}{2}\rho_{N,\omega}^{2}, \tfrac{\theta^{2}}2(C_{3,\min} -1)^{2} + \tfrac{\sigma^{2}}2C_{3,\min}^{2} \Big\}\ .
  \end{align*}
    For $k=n$ we have 
  \begin{align*}
    \min_{\T \in \cB_{N,k,\omega}}\cE_{\textup{i.i.d.}}(\T) &= \tfrac{\mu^{2}}{2}n(C_{2, \min}-1)^{2} + \tfrac{\theta^{2}}{2}(C_{2, \min}-1)^{2}\\ 
    &\quad +  \tfrac{\theta^{2}}{2}(n-1)(C_{3, \min}-1)^{2} + \tfrac{\sigma^{2}}{2}C_{2, \min}^{2} + \tfrac{\sigma^{2}}{2}(n-1)C_{3, \min}^{2}\ .
  \end{align*}
  The minimal risk is attained in both cases. 
  For $N$ odd and $k<n$ it holds for the minimal risk 
  \begin{alignat*}{2}
   \min_{\T\in\cB_{N,k,\omega}} \cE_{\textup{i.i.d.}}(\T) &= \tfrac{\theta^{2}}{2}(k-1)(C_{3,\max} - 1)^{2} + &&\tfrac{\theta^{2}}{2}(n-k)(\rho_{N,\omega}-1)^{2} + \tfrac{\sigma^{2}}{2}(k-1)C_{3,\max}^{2}\\ 
   &\quad + \tfrac{\sigma^{2}}{2}(n-k)\rho_{N,\omega}^{2} + \min\Big\{&&\tfrac{\mu^{2}}{2}n(C_{2,\max} -1)^{2} + \tfrac{\theta^{2}}{2}(C_{2,\max} -1)^{2} + \tfrac{\sigma^{2}}{2}C_{2,\max}^{2},\\
      &\quad &&\tfrac{\mu^{2}}{2}n(\rho_{N,\omega}-1)^{2} + \tfrac{\theta^{2}}2(C_{3,\max} -1)^{2} + \tfrac{\sigma^{2}}2C_{3,\max}^{2} \Big\}\ .
  \end{alignat*}
  For $k=n$ we have 
    \begin{align*}
    \min_{\T \in \cB_{N,k,\omega}}\cE_{\textup{i.i.d.}}(\T) &= \tfrac{\mu^{2}}{2}n(C_{2, \max}-1)^{2} + \tfrac{\theta^{2}}{2}(C_{2, \max}-1)^{2}\\ 
    &\quad +  \tfrac{\theta^{2}}{2}(n-1)(C_{3, \max}-1)^{2} + \tfrac{\sigma^{2}}{2}C_{2, \max}^{2} + \tfrac{\sigma^{2}}{2}(n-1)C_{3, \max}^{2}\ .
  \end{align*}
  Again, in both cases the minimal risk is attained. 
\end{theorem}

The proof is a long computation along the same lines as the proof of Theorem~\ref{thm:best-unrolling-constant}
 and can be found in \ref{app:further-proofs}.

\begin{remark}\label{rem:optimize-omega}
  All the risks in the unrolling case depend on the stepsize $\omega$ and it seems desirable to compute the best $\omega$ and the respective best risks in all cases. These computations are quite cumbersome due the complicated expressions and case distinctions. However, we performed some of the computations in \ref{sec:appendix}. 
  One result that is, in our opinion, remarkable is that the best risk is independent of the number of iterations $N$ in the even case for both data models. In the odd case we could not explicitly optimize over $\omega$, since the risks depend on the quantity $c_{N,\omega}$, which cannot be computed explicitly. In Section~\ref{sec:discussion-visualization} we use numerical optimization to show results for optimized $\omega$.
\end{remark}

\section{Visualization}
\label{sec:discussion-visualization}
In this section we provide some visualizations to illustrate the results from previous sections and give an experiment with real world data. The section is divided into three parts. The first subsection contains a relative comparison of the best risks of each approach with the best linear risk and also with each other. In the second subsection we compare the two approaches in an experiment with real world data and observe some, but not all, of the theoretically proven results. In the third part we give possible explanations for the differences in theory and practice.

\subsection{Best risks}
\label{sec:visualization_of_best_risks}
We start with comparing the risk of each approach with the corresponding best linear risk. Since we are not interested in the actual values of the risks, we compare \emph{risk ratios}, i.e., in this case, quotients of the best linear risk and the risk of some other estimator. The risk ratio of the best linear risk and the lower bound of the bilevel risk, for example, is given by 
\begin{align*}
  \text{Risk Ratio} = \tfrac{\cE_{\text{linear}}}{\cE_{\text{bilevel}}}
\end{align*}
and by construction it holds that $0< \text{Risk Ratio}\leq 1$. The considered risk ratios are explicitly mentioned in each figure. 
To reduce the number of variable parameters, we can normalize the square of the expected value of the signal $\mu^{2}$ to one. We do not lose any generality, since we could just divide the risks in Corollary~\ref{cor:true-risks} by $\mu^{2}$ and would get
\begin{align*}
  \cE_{\textup{const.}} &= \tfrac{1+ \tfrac{\theta^{2}}{\mu^{2}}}2\norm{(\T-\I)\1}_{2}^{2} + \tfrac{\sigma^{2}}{2\mu^{2}}\norm{\T}_{F}^{2}, \,\text{ and }\\
  \cE_{\textup{i.i.d.}} &= \tfrac12\norm{(\T-\I)\1}_{2}^{2} + \tfrac{\theta^{2}}{2\mu^{2}}\norm{\T-\I}_{F}^{2}+\tfrac{\sigma^{2}}{2\mu^{2}}\norm{\T}_{F}^{2},
\end{align*}
which are exactly the risks with $\mu=1$ and scaled versions of $\theta$ and $\sigma$.
Moreover, we fix $n = 500$ and consider $k = 1,\ldots, n$.

\begin{figure}
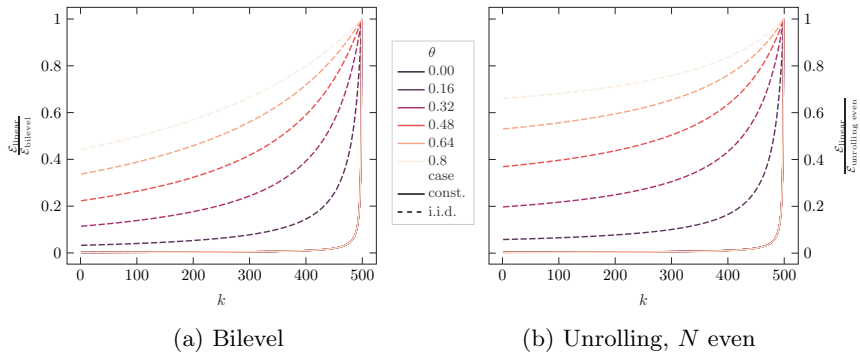

  \centering
  \captionsetup[subfigure]{skip=6pt,margin={0pt,10pt}}
  \subcaptionbox{Bilevel\label{fig:ratio_linear_bilevel}}{\scalebox{0.6}{\input{plots/plot_ratio_linear_bilevel_k}}}
  \captionsetup[subfigure]{skip=6pt,margin={0pt,30pt}}
  \subcaptionbox{Unrolling, $N$ even\label{fig:ratio_linear_unrolling_even}}{\scalebox{0.6}{\input{plots/plot_ratio_linear_unrolling_even_k}}}
  \captionsetup{subrefformat=parens}
  \caption{Plots of the risk ratios of the best linear risk and the lower bound of the bilevel or the unrolling risk for $N$ even for given values $n=500$, $\mu=1.0$ and $\sigma=0.9$.}
\end{figure}

   In Figure~\ref{fig:ratio_linear_bilevel} we show the risk ratio of the best linear risk and the bilevel risk for both data models and the given values of the parameters. The risk ratio is generally much higher for the i.i.d. model than for the model of constant random vectors meaning that the bilevel estimations are closer to the best linear estimators in this case. In all cases, increasing $k$ brings the performance of the bilevel method closer to the best linear model. This is to be expected, since Theorem~\ref{thm:expressivity-bilevel} shows that larger $k$ leads to a more expressive model. Moreover, larger values for the signal variance $\theta^{2}$ lead to a smaller gap in the risks. Note that this is also true for the constant vectors data model, even though their lines seem to lie on top of each other.

   In the case of unrolling we have two more parameters: The number of iterations $N$ and the stepsize $\omega$. 
   We show individual figures for an even and an odd case, since they behave very differently. Moreover, we show the risks for optimal $\omega$, i.e., after the expressions have been optimized with respect to $\omega$  (see \ref{sec:appendix}).

\begin{figure}
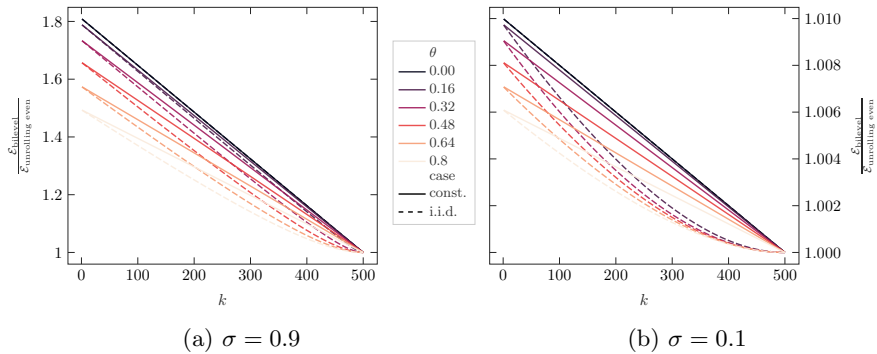

  \centering
  \captionsetup[subfigure]{skip=6pt,margin={0pt,0pt}}
  \subcaptionbox{$\sigma=0.9$\label{fig:ratio_linear_unrolling_bilevel_0.9}}{\scalebox{0.6}{\input{plots/plot_bilevel_unrolling_even_k_0.9}}}
  \captionsetup[subfigure]{skip=6pt,margin={0pt,0pt}}
  \subcaptionbox{$\sigma=0.1$\label{fig:ratio_linear_unrolling_bilevel_0.1}}{\scalebox{0.6}{\input{plots/plot_bilevel_unrolling_even_k_0.1}}}
  \captionsetup{subrefformat=parens}
  \caption{The ratios between the lower bound of the bilevel risk and the unrolling risk for $N$ even for given values $n=500$, $\mu=1.0$ and different values of $\sigma$. }\label{fig:ratio_bilevel_unrolling_even_k}
\end{figure}

   Figure~\ref{fig:ratio_linear_unrolling_even} shows the risk ratio for the case of even $N$. Since, by Remark~\ref{rem:optimize-omega}, the risk is independent of the number $N$ of unrolled iterations if one optimizes over the stepsize $\omega$ (see \ref{sec:appendix} for details), we do not show the dependence on $N$ in these plots.
   Overall Figure~\ref{fig:ratio_linear_unrolling_even} looks quite similar to Figure~\ref{fig:ratio_linear_bilevel}. In case of the model of random constant vectors the plots only differ slightly (see Figure~\ref{fig:ratio_bilevel_unrolling_even_k} below for another visualization), but for the i.i.d. model the risk ratio for unrolling is much higher than for bilevel. Moreover, the difference between the risk ratios for unrolling and bilevel increases with higher values of $\theta$.

\begin{figure}
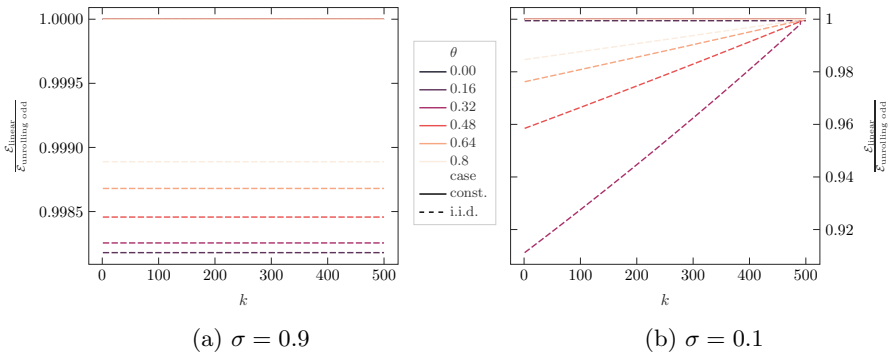

  \centering
  \captionsetup[subfigure]{skip=6pt,margin={0pt,0pt}}
  \subcaptionbox{$\sigma=0.9$\label{fig:ratio_linear_unrolling_odd_0.9}}{\scalebox{0.6}{\input{plots/plot_ratio_linear_unrolling_odd_k_0.9}}}
  \captionsetup[subfigure]{skip=6pt,margin={0pt,0pt}}
  \subcaptionbox{$\sigma=0.1$\label{fig:ratio_linear_unrolling_odd_0.1}}{\scalebox{0.6}{\input{plots/plot_ratio_linear_unrolling_odd_k_0.1}}}
  \captionsetup{subrefformat=parens}
  \caption{The ratios between the best linear risk and the unrolling risk for $N$ odd for given values $N = 5$, $n=500$, $\mu=1.0$ and different values of $\sigma$.} \label{fig:ratio_linear_unrolling_odd}
\end{figure}
   In Figure~\ref{fig:ratio_linear_unrolling_odd} we compare the best linear risks with the unrolling risk in the odd case $N=5$ for both data models and the given values of the parameters. To compute the risks in this case $c_{N,\omega}$ from Theorem~\ref{thm:expressivity-unrolling} is needed and we use numerical optimization to compute this value. Considering only Figure~\ref{fig:ratio_linear_unrolling_odd_0.9} which corresponds to $\sigma = 0.9$, it seems to show that the ratio is independent of $k$. However, the plot for $\sigma = 0.1$ Figure~\ref{fig:ratio_linear_unrolling_odd_0.1} shows that the risk ratio in the i.i.d. case increases with increasing $k$.

  Since Figure~\ref{fig:ratio_linear_bilevel} and Figure~\ref{fig:ratio_linear_unrolling_even} look quite similar, we show the risk ratio of the bilevel method and the unrolling method, i.e.,
  \begin{align*}
    \tfrac{\cE_{\text{bilevel}}}{\cE_{\text{unrolling even}}},
  \end{align*}   
  for the case of even $N$ for $\sigma = 0.9$ and $\sigma=0.1$ in Figure~\ref{fig:ratio_bilevel_unrolling_even_k}. Note that in this case $0< \text{Risk Ratio}\leq 1$ does not hold anymore and values greater one are possible depending on the performance of the two compared approaches. Notably, the risk ratios are always greater than one, meaning that the unrolling risk is always lower.
  Moreover it can be seen that the ratio between the lower bound of the bilevel risk and the unrolling even risk decreases with increasing $k$. Considering different values of $\sigma$, smaller values lead to a smaller gap in the risks for unrolling and bilevel. Moreover, larger values for $\theta$ lead to a smaller gap in the risks. 

\subsection{Experiment on real world data}
\label{sec:visualization_part_2}
We conducted an experiment with speech data from the IEEE speech corpus \cite{loizou2013speech}. Ground truth vectors of length $n=320$, extracted from the speech signals, are contaminated by additive Gaussian noise with zero mean and standard deviation $\sigma=0.1$.
   \begin{figure}[htb]
    \begin{center}
    \input{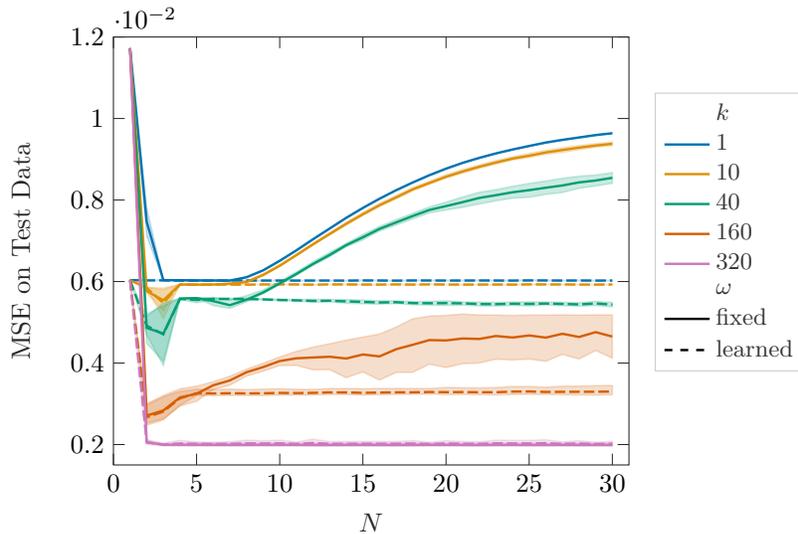}\caption{Empirical risks for learned and fixed stepsizes in dependence of $N$.}\label{fig:experiment_risk}
    \end{center}
   \end{figure}
   \begin{figure}[htb]
    \begin{center}
\begin{tikzpicture}

\definecolor{chocolate213940}{RGB}{213,94,0}
\definecolor{darkcyan1115178}{RGB}{1,115,178}
\definecolor{darkcyan2158115}{RGB}{2,158,115}
\definecolor{darkorange2221435}{RGB}{222,143,5}
\definecolor{darkslategray38}{RGB}{38,38,38}
\definecolor{lightgray204}{RGB}{204,204,204}
\definecolor{orchid204120188}{RGB}{204,120,188}

\begin{axis}[
axis line style={white!15!black},
legend cell align={left},
legend style={
  font=\small, 
  fill opacity=0.8,
  draw opacity=1,
  text opacity=1,
  at={(1.05,0.87)},
  anchor=north west,
  draw=white!80!black
},
tick align=inside,
x grid style={white!80!black},
xlabel=\textcolor{white!15!black}{\(\displaystyle N\)},
xmajorticks,
xmin=0, xmax=31,
xtick style={color=white!15!black},
y grid style={white!80!black},
ylabel=\textcolor{white!15!black}{Best unrolling risk from Theorem~\ref{thm:best-unrolling-constant}},
ymajorticks,
ymin=0.0015, ymax=0.012,
ytick style={color=white!15!black}
]
\addlegendimage{empty legend}
\addlegendentry{$k$}
\newcommand{\om}{0.1269} 
\newcommand{\thet}{0.00004}
\newcommand{\mue}{0.00004}
\newcommand{\sig}{0.00006}

\addplot[domain=0:31,very thick,color=darkcyan1115178] {(\mue + \thet)/2*320*(1-\om)^(2*x) + \sig/2*(320-1)*(1-(1-\om)^x)^2};
\addlegendentry{$1$}
\addplot[domain=0:31,very thick,color=darkorange2221435] {(\mue + \thet)/2*320*(1-\om)^(2*x) + \sig/2*(320-10)*(1-(1-\om)^x)^2};
\addlegendentry{$10$}
\addplot[domain=0:31,very thick,color = darkcyan2158115] {(\mue + \thet)/2*320*(1-\om)^(2*x) + \sig/2*(320-40)*(1-(1-\om)^x)^2};
\addlegendentry{$40$}

\addplot[domain=0:31,very thick,color=chocolate213940] {(\mue + \thet)/2*320*(1-\om)^(2*x) + \sig/2*(320-160)*(1-(1-\om)^x)^2};
\addlegendentry{$160$}
\addplot[domain=0:31,very thick,color=orchid204120188] {(\mue + \thet)/2*320*(1-\om)^(2*x)};
\addlegendentry{$320$}

\addlegendimage{empty legend}

\end{axis}

\end{tikzpicture}\caption{Best risks according to Theorem~\ref{thm:best-unrolling-constant} for some chosen values of $\theta,\mu$ and $\sigma$ and $\omega = \log(1+\exp(-2))$ (similar to fixed stepsize $\omega$ in Figure~\ref{fig:experiment_risk}) in dependence on $N$.}\label{fig:compare_experiment_risk}
    \end{center}
   \end{figure}
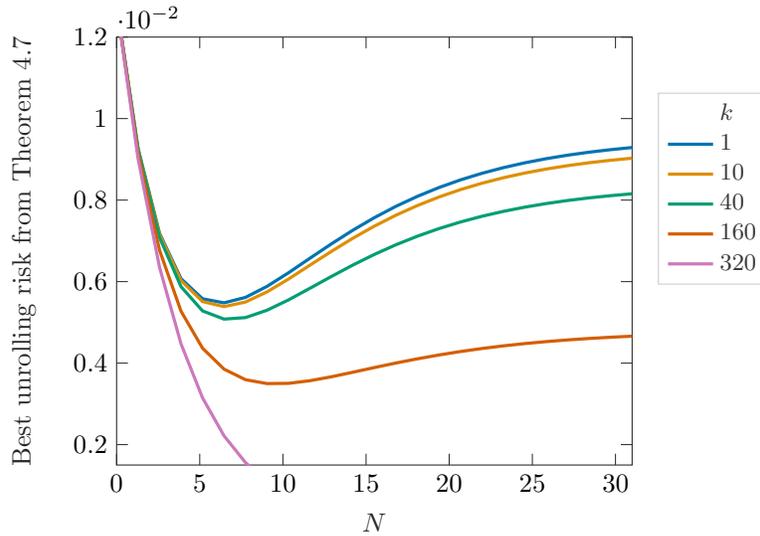
   \begin{figure}[htb]
    \begin{center}
    \input{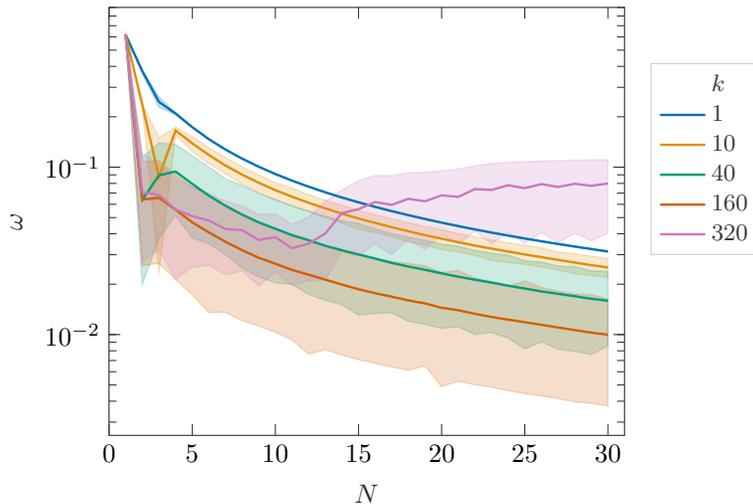}\caption{Semilogplot of the learned stepsizes in dependence of $N$.}\label{fig:experiment_stepsizes}
    \end{center}
   \end{figure}

   In Figure \ref{fig:experiment_risk} we show the mean squared error, i.e., the empirical risk, for different values of $k$ in dependence of the number of iterations $N$. Moreover, we distinguish between fixed stepsizes and stepsizes which are learned during training. At first, all risks decrease with increasing $N$. While for fixed $\omega$ the risks starts to increase after some iterations, for learned stepsizes the risk does not change after a few iterations. This has been proven for the even case as after optimizing over the stepsize the optimal risk is independent of $\omega$. For fixed $k$, the risk with learned stepsizes is at least not worse than for fixed stepsizes and better in most cases. Moreover, increasing $k$ results in lower risks in all cases.

   For a comparison we show in Figure~\ref{fig:compare_experiment_risk} the best risks of the unrolling estimator in the case of random constant vectors from Theorem~\ref{thm:best-unrolling-constant}. We took $\omega$ exactly as in the numerical experiment with fixed stepsize and also for $n$ and $k$ we chose the same values. The values of $\theta,\mu$ and $\sigma$ have been adapted to bring the plots in the same order of magnitude, but we stress that the general shape of the risk (first decreasing and then increasing again) is independent of the choice of these parameters. As can be seen from Figure~\ref{fig:compare_experiment_risk}, the increase of the MSE for increasing depth $N$ is indeed predicted by our results and is not an artifact of the training process. Hence, the worse risks for deeper unrolling are inherent to the unrolling approach and not just due to insufficient training.

    Figure \ref{fig:experiment_stepsizes} shows the learned stepsizes in dependence of the number of iterations $N$. The learned stepsizes decrease with increasing $N$ except for $k=n=320$. In that case, the stepsize starts increasing after it decreased for a few iterations. Also, larger $k$ results in smaller learned stepsizes (except for the special case $k=n$). Note, that decreasing stepsizes even in regimes of $N$ in which the risk does not change anymore are not a contradiction to the theory since the risk is independent of $N$ but $\omega$ still depends on it.
  
\subsection{Discrepancy in observations}
\label{sec:visualization_part_3}
While some of the behaviour in the plots of the experiments is according to the theoretical results, there are observations which are not following or even contradictory to the theory. Here, we will give possible explanations for those. While investigating the plots in Figures~\ref{fig:experiment_risk} and \ref{fig:experiment_stepsizes} one notices two claims from theory which are not confirmed from the experiment: the parity of $N$ seems to not make a difference and increasing $k$ results in different behaviour of the learned stepsizes than theory claims for optimal stepsizes.
 
To see why the difference in the risks or the stepsizes for an odd or an even number of iterations is not seen in the experiments we consider the case of $n=k=1$ which corresponds to $\RRR=r$ being a scalar. Then, the unrolling estimator is also a scalar
\begin{equation*}
  \T=T=\tfrac{1-(1-\omega(1+r^{2}))^{N}}{1+r^{2}}
\end{equation*}
and we get the risk
\begin{align*}
  \cE_{\textup{const.}}(T) & = \tfrac{\mu^{2}+ \theta^{2}}2\abs{T-1}^{2} + \tfrac{\sigma^{2}}2\abs{T}^{2}\\
  &  = \tfrac{\mu^{2}}2\abs{T-1}^{2} + \tfrac{\theta^{2}}2\abs{T-1}^{2}+\tfrac{\sigma^{2}}2\abs{T}^{2} = \cE_{\textup{i.i.d.}}(T)
\end{align*}
(especially the two situations of random constant vectors and random i.i.d. vectors result in the same risk).

\begin{figure}[htb]
  \begin{center}
    \input{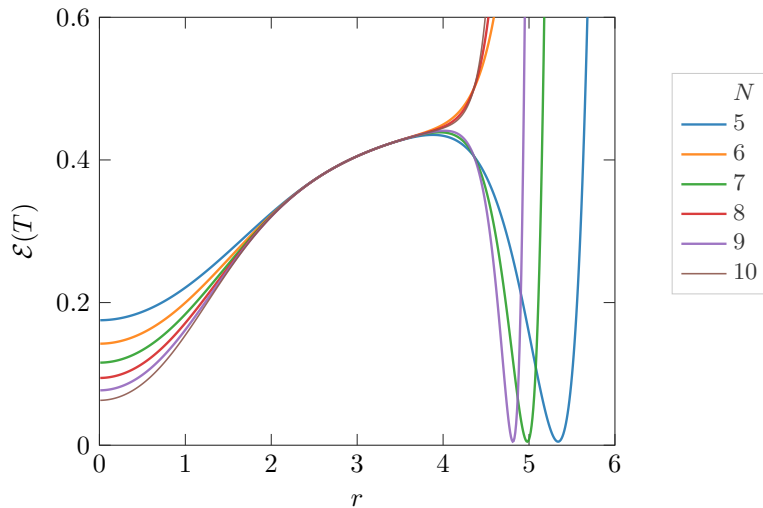}\caption{The risk for $\omega=0.1$, $n=k=1$, $\mu=1.0$, $\sigma=0.1$, $\theta=0.02$ and different values (even and odd) of $N$.}\label{fig:risk_k_n_1}
  \end{center}
\end{figure}

The plot in Figure~\ref{fig:risk_k_n_1} shows the true risk for multiple even and odd $N$ with a fixed, not optimized $\omega$. It can be seen that for small $r$, the graphs are fairly close (they all have a local minimum at a small but positive $r$) and this changes at around $r=4$. While for even $N$ the risk already diverges, there exists a second, slightly smaller local minimum for odd $N$ around $r=5$. Those second local minima are very steep and hence, not easily reached by optimization methods in practical applications. Moreover, the steepness increases with increasing $N$ which worsens the situation. Those steep local minima may lead to the fact that the theoretically proven difference in unrolling for $N$ even and $N$ odd do not appear in practical applications, as in practice, one only sees the local minima for small $r$, and these are quite close for $N$ even and odd, respectively. Nevertheless, we still have the results in Theorem~\ref{thm:expressivity-unrolling} proving different expressivity depending on the parity of $N$.

While the behaviour of the risk for increasing $k$ is as shown in the theory (decreasing risk with increasing $k$), the learned stepsizes in Figure~\ref{fig:experiment_stepsizes} do not behave as the theoretical optimal stepsizes do. Theory claims, that, with increasing $k$, the optimal stepsizes increase. In Figure~\ref{fig:experiment_stepsizes} it is shown that the learned stepsizes decrease with increasing $k$. This, again, supports the conjecture that (some) optimal risks along with the optimal stepsizes are not reached during the training.

\section{Discussion}
\label{sec:discussion}

We derived best risks and estimators in the case of linear, bilevel and unrolling estimators for both our data models (some estimators have complicated forms and can only be found in \ref{app:further-proofs}). One simple observation is that the best linear estimator from Section~\ref{sec:best-linear} is a rank-one matrix in the case of the random constant model and a convex combination of the identity and a rank-one matrix in the case of the i.i.d. vectors.  By our expressivity results from Section~\ref{sec:expressivity} we see that especially the bilevel approach can never reach the rank-one estimator as all bilevel estimators are always of full rank, and for $k=n-1$ we can only approximate the optimal estimator since the lower bound on the eigenvalues in Theorem~\ref{thm:expressivity-bilevel} is strict. The unrolling estimators, described in Theorem~\ref{thm:expressivity-unrolling}, behave differently: For even $N$ and $0<\omega<2$ we still have estimators of rank $n-k$ and we can get a rank-one estimator for $k=n-1$, but our results from the previous sections show that the optimal estimators are not reached by unrolling. For odd $N$ we always have a positive lower bound on all eigenvalues, so the rank-one estimator can never be reached in this case.

Next we discuss our results and approach in the light of real world applications, and the limitations of our toy model.
The models for noise and data that we defined in Section~\ref{sec:model} are indeed very simple. However, the general problem of denoising is widely used as a toy model for more complicated models. Moreover, learned denoising is still an active area of research \cite{tian2020deep,zheng2021deep}, although in more complicated scenarios than considered here.

While our noise model (additive white noise) is quite common (and also realistic in several applications), the data model is extremely simple: The signals are either constant (random constant) or a slightly perturbed constant (i.i.d. with constant mean and variance). Random constant vectors are a meaningful first step for the more realistic model of piecewise constant vectors or even vectors of the form $y = \sum_i \alpha_{i}\phi_{i}$ with a known and fixed basis $\phi_{i}$ and random coefficients $\alpha_{i}$. The second model of i.i.d. vectors has been previously used for theoretical studies of risks of regression models and is known as ``random-effects model'', see, e.g.,\cite{dobriban2018high,mucke2022data}.

Our objective consists of a quadratic discrepancy and a quadratic penalty. The quadratic discrepancy can be derived as negative log-likelihood of the noise distribution (compare~\cite[Section 3.1]{kaipio2006statistical}) but the quadratic penalty is not justified by our data models and is used only for simplicity. However, penalties of the form $\sum \theta_{i}\rho_{i}(\RRR_{i}\xx)$ with weights $\theta_{i}$, real functions $\rho_{i}$ and matrices $\RRR_{i}$ (sometimes known as \emph{field of experts}~\cite{roth2005fields,Roth2009}) are common. Our model is the special case where the $\RRR_{i}$ are row vectors, $\rho_{i}(t) = t^{2}$, $\theta_{i}=1$ for all $i$. Moreover, quadratic penalties also appear as subproblems in iteratively reweighted algorithms~\cite{vu2021}. A paper that studies unrolling in a quite similar model as we do is~\cite{le2022faster}, where accelerated proximal algorithms are unrolled and mainly examined experimentally.

Finally, we considered a very simple algorithm to unroll (gradient descent) for which we could write down all iterates in a fairly simple closed form. However, gradient descent is a special case of the more general class of proximal gradient methods and hence, quite realistic in our scenario. While the extension of our results to proximal gradient methods would require a totally different analysis, the extension to the more elaborate algorithms like accelerated gradient descent could, in principle, be done along the lines of the presented work. Moreover, gradient descent with constant stepsize is exactly the Landweber method~\cite{landweber1951iteration}. This method, applied to just the least squares data-fit term, is used as regularization method for ill-posed inverse problems~\cite{engl1996regularization} and it is indeed a regularization when combined with an appropriate early-stopping technique. In our context this is exactly unrolling of gradient descent as a replacement for the exact solution of the lower level problem.

\section{Conclusion and outlook}
\label{sec:conclusion}

In this paper we started a systematic investigation of the effect of algorithm unrolling for learning of variational models. Our approach was to consider a simple example where everything can be computed (almost) explicitly. Even this simple example illustrates that some phenomena that have been observed in practice can be explained theoretically, e.g. that learning the stepsize has a big impact while the number of unrolled iterations may have very different effects. Our analysis relied on the true risk and a natural next step is to extend this analysis to empirical risks. Moreover, more general inverse problems could be considered, i.e., we could replace the lower level problem by $\min_{\z} \tfrac12\norm{\A\z-\xx}_{2}^{2} + \tfrac12\norm{\RRR\xx}_{2}^{2}$~\cite{alberti2021learning} as well as more general regularizers like, e.g. fields of experts regularizers of the form $\sum_{i}\theta_{i}\rho(\RRR_{i}\xx)$ with scalar weights $\theta_{i}$, matrices $\RRR_{i}\in\R^{n_{i}\times n}$ and functions $\rho:\R^{n_{i}}\to \R$~\cite{Chen2014,Roth2009}. However, both these generalizations would need new technical approaches.

\appendix

\section{Optimized $\omega$ for Unrolling}
\label{sec:appendix}
  We give the optimal $\omega$ and its corresponding risk for unrolling in the setting of $N$ even. Since $c_{N,\omega}$ is not known explicitly, we investigate the situation of $N$ odd numerically. Plots for both scenarios are given in the visualization part. \\
  
   \changes{
  \begin{figure}
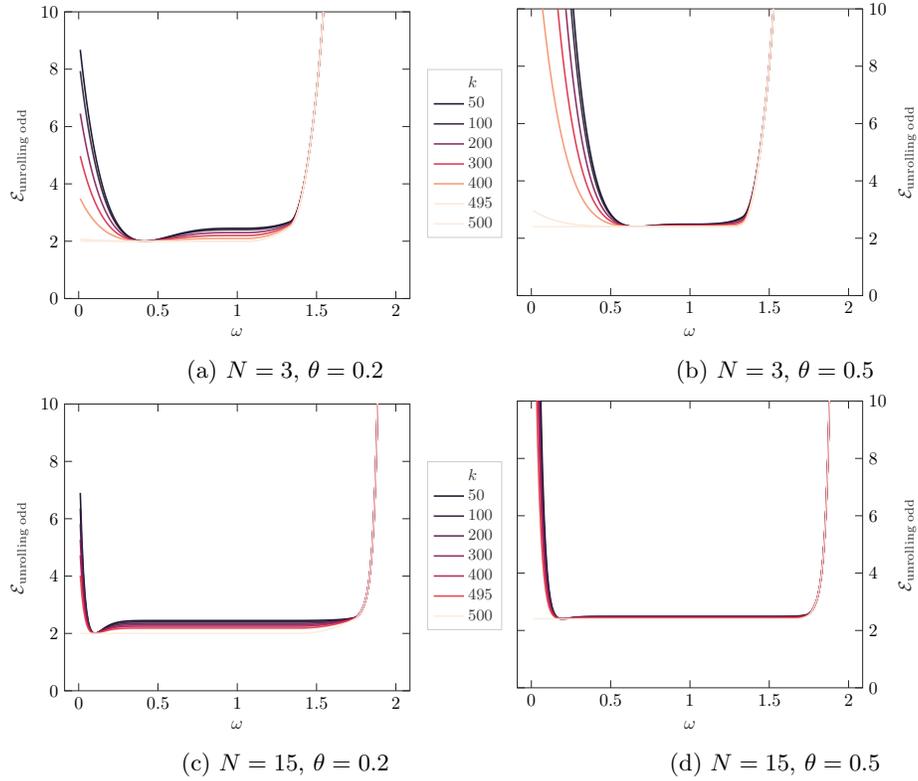

    \centering
    \captionsetup[subfigure]{skip=6pt,margin={25pt,0pt}}
     \subcaptionbox{$N=3$, $\theta=0.2$\label{fig:plot_risk_unrolling_odd_omega_3_0.2}}{
    \scalebox{0.67}{\input{plots/plot_risk_unrolling_odd_omega_3}}}
   \captionsetup[subfigure]{skip=6pt,margin={40pt,0pt}}
     \subcaptionbox{$N=3$, $\theta=0.5$\label{fig:plot_risk_unrolling_odd_omega_3_0.5}}{\scalebox{0.67}{\input{plots/plot_risk_unrolling_odd_omega_3_0.5}}}
    \centering
    \captionsetup[subfigure]{skip=6pt,margin={25pt,0pt}}
     \subcaptionbox{$N=15$, $\theta=0.2$\label{fig:plot_risk_unrolling_odd_omega_15_0.2}}{
    \scalebox{0.67}{\input{plots/plot_risk_unrolling_odd_omega_15}}}
   \captionsetup[subfigure]{skip=6pt,margin={40pt,0pt}}
     \subcaptionbox{$N=15$, $\theta=0.5$\label{fig:plot_risk_unrolling_odd_omega_15_0.5}}{\scalebox{0.67}{\input{plots/plot_risk_unrolling_odd_omega_15_0.5}}}
     \captionsetup{subrefformat=parens}
     \caption{The unrolling risks for an odd number of iterations dependent of $\omega$ for given values $n=500$, $\mu=1.0$, $\sigma=0.1$ and different values of $N$ and $\theta$ in the random i.i.d. model. }\label{fig:plot_risk_unrolling_odd_omega_15}
  \end{figure}
Before we start the analysis of optimal $\omega$ we show the unrolling risk for an odd number of iterations in the i.i.d. model in dependence of the stepsize $\omega$ for $N=3$ and $N=15$, respectively, in Figure~\ref{fig:plot_risk_unrolling_odd_omega_15}. Even though the graphs are not the same, it can be seen that the optimal values for the risks are independent not only from $k$ but also from the number of unrolled steps $N$. However, larger values of $\theta$ lead to a more flat region close to the optimal $\omega$ which may lead to inaccuracies in the approximation of $c_{N,\omega}$ and, thus, calculating the optimal $\omega$. Moreover, it can be seen that larger values of $\theta$ correspond to larger values of the optimal risk.
}

  \textbf{Random constant vectors} 
  \begin{itemize}
    \item $N$ even: \\
    For $k<n$ it is  
    \begin{align*}
      \min_{\T\in\cB_{N,k,\omega}} \cE_{\textup{const.}}(\T) &= \tfrac{\mu^{2} + \theta^{2}}{2}n(\rho_{N,\omega} - 1)^{2} + \tfrac{\sigma^{2}}{2}(n-k)\rho_{N,\omega}^{2} \\
      &= \tfrac{\mu^{2} + \theta^{2}}{2}n(1-\omega)^{2N} + \tfrac{\sigma^{2}}{2}(n-k)(1-(1-\omega)^{N})^{2} \ .
    \end{align*}
    A differentiation with respect to the stepsize $\omega$ gives 
    \begin{align*}
      -2Nn\tfrac{\mu^{2} + \theta^{2}}{2}(1-\omega)^{2N-1} + 2N \tfrac{\sigma^{2}}{2}(n-k)(1-(1-\omega)^{N})(1-\omega)^{N-1} = 0
    \end{align*}
    which leads to optimal $\omega^{*} = 1 \pm \sqrt[N]{\tfrac{\sigma^{2}(n-k)}{n(\mu^{2} + \theta^{2}) + \sigma^{2}(n-k)}}$. The corresponding risk is 
    \begin{align*}
      \cE_{const., k<n}^{*} = \tfrac12\tfrac{n(n-k)(\mu^{2} + \theta^{2})\sigma^{2}}{n(\mu^{2} + \theta^{2}) + (n-k)\sigma^{2}}.
    \end{align*}
    For $k = n$ we have 
    \begin{align*}
          \min_{\T \in \cB_{N,k,\omega}}\cE_{\textup{const.}}(\T) = \tfrac{\mu^{2} + \theta^{2}}{2}n(C_{1, \min}-1)^{2} + \tfrac{\sigma^{2}}{2}C_{1, \min}^{2} \ . 
    \end{align*}
    with $C_{1,\min} = \min\{\tfrac{(\mu^{2} + \theta^{2})n}{(\mu^{2} + \theta^{2})n + \sigma^{2}}, \rho_{N,\omega}\}$. A case distinction in $C_{1,\min}$ and further transformations lead to 
     \begin{align*}
          \min_{\T \in \cB_{N,k,\omega}}\cE_{\textup{const.}}(\T) = \begin{cases}
            \tfrac{\mu^{2} + \theta^{2}}{2}n(\tfrac{(\mu^{2} + \theta^{2})n}{(\mu^{2} + \theta^{2})n + \sigma^{2}}-1)^{2} + \tfrac{\sigma^{2}}{2}\tfrac{(\mu^{2} + \theta^{2})n}{(\mu^{2} + \theta^{2})n + \sigma^{2}}^{2}&, \qquad \text{if } \omega\in \mathcal{I}\\
            \tfrac{\mu^{2} + \theta^{2}}{2}n(\rho_{N,\omega} - 1)^{2} + \tfrac{\sigma^{2}}{2}(n-k)\rho_{N,\omega}^{2}&, \qquad \text{else}
            \end{cases}
    \end{align*}
    with $\mathcal{I}:= \Big]1-\sqrt[N]{\tfrac{\sigma^{2}}{(\mu^{2} + \theta^{2})n + \sigma^{2}}},1+\sqrt[N]{\tfrac{\sigma^{2}}{(\mu^{2} + \theta^{2})n + \sigma^{2}}}\Big[$. An optimization over $\omega$ for $\omega \notin \mathcal{I}$ gives the optimal stepsize $\omega^{*}=1\pm \sqrt[N]{\tfrac{\sigma^{2}}{(\mu^{2} + \theta^{2})n + \sigma^{2}}}$ with corresponding risk 
    \begin{align*}
       \tfrac{\mu^{2} + \theta^{2}}{2}n(\tfrac{(\mu^{2} + \theta^{2})n}{(\mu^{2} + \theta^{2})n + \sigma^{2}}-1)^{2} + \tfrac{\sigma^{2}}{2}\tfrac{(\mu^{2} + \theta^{2})n}{(\mu^{2} + \theta^{2})n + \sigma^{2}}^{2}\ .
    \end{align*}
    Since this is equal to the risk for $\omega \in \mathcal{I}$ (in which no $\omega$ appears), the set of optimal stepsizes is given by $[1 - \sqrt[N]{\tfrac{\sigma^{2}}{(\mu^{2} + \theta^{2})n + \sigma^{2}}},1 + \sqrt[N]{\tfrac{\sigma^{2}}{(\mu^{2} + \theta^{2})n + \sigma^{2}}}]$ and the corresponding optimal risk is 
    \begin{align*}
      \cE_{const., k=n}^{*} &= \tfrac{\mu^{2} + \theta^{2}}{2}n(\tfrac{(\mu^{2} + \theta^{2})n}{(\mu^{2} + \theta^{2})n + \sigma^{2}}-1)^{2} + \tfrac{\sigma^{2}}{2}\tfrac{(\mu^{2} + \theta^{2})n}{(\mu^{2} + \theta^{2})n + \sigma^{2}}^{2}\\
      &=      \tfrac{1}{2}\tfrac{n(\mu^{2} + \theta^{2})\sigma^{2}}{n(\mu^{2} + \theta^{2}) + \sigma^{2}} = \cE_{const., k=n-1}^{*}\ .
    \end{align*}
    \item $N$ odd: \\
    Here, we do an numerical investigation of the risk optimized over $\omega$ since there are only upper and lower bounds for $c_{N,\omega}$, which appears in the explicit formulas of the risk.
  \end{itemize}
\textbf{Random i.i.d. vectors}\\
\begin{itemize}
  \item $N$ even: \\
  For $k<n$ it is 
    \begin{align*}
    \min_{\T\in\cB_{N,k,\omega}} \cE_{\textup{i.i.d.}}(\T) &= \tfrac{\mu^{2}}{2}n(\rho_{N,\omega} -1)^{2} + \tfrac{\theta^{2}}{2}(k-1)(C_{3,\min} - 1)^{2} + \tfrac{\theta^{2}}{2}(n-k)(\rho_{N,\omega}-1)^{2}\\
    &\quad + \tfrac{\sigma^{2}}{2}(k-1)C_{3,\min}^{2} + \tfrac{\sigma^{2}}{2}(n-k)\rho_{N,\omega}^{2}\\ 
    &\quad + \min\Big\{\tfrac{\theta^{2}}{2}(\rho_{N,\omega} -1)^{2} + \tfrac{\sigma^{2}}{2}\rho_{N,\omega}^{2}, \tfrac{\theta^{2}}2(C_{3,\min} -1)^{2} + \tfrac{\sigma^{2}}2C_{3,\min}^{2} \Big\}\ .
  \end{align*}
  We consider a case distinction in $C_{3,\min} = \min\{\tfrac{\theta^{2}}{\theta^{2} + \sigma^{2}}, \rho_{N,\omega}\}$ and start with $C_{3,\min} = \rho_{N,\omega}$. Plugging in $C_{3,\min}$ and $\rho_{N,\omega}$ we get 
  \begin{align}\label{eq:optimal-omega-iid-N-even}
    \min_{\T\in\cB_{N,k,\omega}} \cE_{\textup{i.i.d.}}(\T) &= \tfrac{\mu^{2}}{2}n(1-\omega)^{2N} + \tfrac{\theta^{2}}{2}n(1-\omega)^{2N} + \tfrac{\sigma^{2}}{2}n(1-(1-\omega)^{N})^{2} \ .
  \end{align}
  An optimization over $\omega \in \Big]0,1-\sqrt[N]{\tfrac{\sigma^{2}}{\theta^{2} + \sigma^{2}}}\Big[\ \bigcup\ \Big]1+\sqrt[N]{\tfrac{\sigma^{2}}{\theta^{2} + \sigma^{2}}},2\Big[$ (which corresponds to $C_{3,\min} = \rho_{N,\omega}$) shows that the infimum of (\ref{eq:optimal-omega-iid-N-even}) is 
  \begin{align*}
    \tfrac{1}{2}n\tfrac{\sigma^{2}}{(\theta^{2} + \sigma^{2})^{2}}\Big(\mu^{2}\sigma^{2} + \theta^{2}\sigma^{2} + \theta^{4}\Big)
  \end{align*}
  with $\arginf_{\omega} \cE_{i.i.d., k<n}(\T) = 1\pm \sqrt[N]{\tfrac{\sigma^{2}}{\theta^{2} + \sigma^{2}}}$, which is not attained. \\
  For $C_{3,\min} = \tfrac{\theta^{2}}{\theta^{2} + \sigma^{2}}$ which corresponds to $1 - \sqrt[N]{\tfrac{\sigma^{2}}{\theta^{2} + \sigma^{2}}}\leq \omega \leq 1 + \sqrt[N]{\tfrac{\sigma^{2}}{\theta^{2} + \sigma^{2}}}$ we have 
  \begin{align*}
    \min_{\T\in\cB_{N,k,\omega}} \cE_{\textup{i.i.d.}}(\T) &= \tfrac{\mu^{2}}{2}n(1-\omega)^{2} + \tfrac{\theta^{2}}{2}(k-1)(\tfrac{\theta^{2}}{\theta^{2} + \sigma^{2}} - 1)^{2} + \tfrac{\theta^{2}}{2}(n-k)(1-\omega)^{2}\\
    &\quad + \tfrac{\sigma^{2}}{2}(k-1)\tfrac{\theta^{4}}{(\theta^{2} + \sigma^{2})^{2}} + \tfrac{\sigma^{2}}{2}(n-k)(1-(1-\omega)^{N})^{2}\\ 
    &\quad + \min\Big\{\tfrac{\theta^{2}}{2}(1-\omega)^{2} + \tfrac{\sigma^{2}}{2}(1-(1-\omega)^{N})^{2}, \tfrac{\sigma^{2}}2\tfrac{\theta^{2}}{\theta^{2} + \sigma^{2}} \Big\}\ .
  \end{align*}
  Since the first term of the minimum is optimal for $\omega = 1 \pm  \sqrt[N]{\tfrac{\sigma^{2}}{\theta^{2} + \sigma^{2}}}$ with optimal value 
  \begin{align*}
    \tfrac{\theta^{2}}2(\tfrac{\theta^{2}}{\theta^{2} + \sigma^{2}} -1)^{2} + \tfrac{\sigma^{2}}2\tfrac{\theta^{4}}{(\theta^{2} + \sigma^{2})^{2}}\ ,
  \end{align*}
  which is equal to the second term of the minimum, the risk reduces to 
  \begin{align*}
    \min_{\T\in\cB_{N,k,\omega}} \cE_{\textup{i.i.d.}}(\T) &= \tfrac{\mu^{2}}{2}n(1-\omega)^{2} + \tfrac{\theta^{2}}{2}k(\tfrac{\theta^{2}}{\theta^{2} + \sigma^{2}} - 1)^{2} + \tfrac{\theta^{2}}{2}(n-k)(1-\omega)^{2}\\
    &\quad + \tfrac{\sigma^{2}}{2}k\tfrac{\theta^{4}}{(\theta^{2} + \sigma^{2})^{2}} + \tfrac{\sigma^{2}}{2}(n-k)(1-(1-\omega)^{N})^{2}\ .
  \end{align*}
  If we optimize over with respect to the constraint $1 - \sqrt[N]{\tfrac{\sigma^{2}}{\theta^{2} + \sigma^{2}}}\leq \omega \leq 1 + \sqrt[N]{\tfrac{\sigma^{2}}{\theta^{2} + \sigma^{2}}}$ we get $\omega^{*} = 1\pm \sqrt[N]{\tfrac{\sigma^{2}(n-k)}{\mu^{2}n + \theta^{2}(n-k) + \sigma^{2}(n-k)}}$ with optimal risk 
  \begin{align*}
    \cE^{*} = \tfrac{k}{2}\tfrac{\theta^{2\sigma^{2}}}{(\theta^{2} + \sigma^{2})} + \tfrac12 \tfrac{\sigma^{2}(n-k)(\mu^{2}n + \theta^{2}(n-k))}{\mu^{2}n + \theta^{2}(n-k) + \sigma^{2}(n-k)}\ .
  \end{align*}
  For $k=n$ we have 
    \begin{align*}
    \min_{\T \in \cB_{N,k,\omega}}\cE_{\textup{i.i.d.}}(\T) &= \tfrac{\mu^{2}}{2}n(C_{2, \min}-1)^{2} + \tfrac{\theta^{2}}{2}(C_{2, \min}-1)^{2}\\ 
    &\quad +  \tfrac{\theta^{2}}{2}(n-1)(C_{3, \min}-1)^{2} + \tfrac{\sigma^{2}}{2}C_{2, \min}^{2} + \tfrac{\sigma^{2}}{2}(n-1)C_{3, \min}^{2}\ .
  \end{align*}
  with $C_{2,\min} = \min\{ \tfrac{\mu^{2}n + \theta^{2}}{\mu^{2}n + \theta^{2} + \sigma^{2}}, \rho_{N,\omega}\}$ and $C_{3,\min}$ as above. Here, we have to consider the three different cases
  \begin{enumerate}
    \item $C_{2,\min} = C_{3,\min} = \rho_{N,\omega}$
    \item $C_{2,\min} = \rho_{N,\omega}, C_{3,\min} = \tfrac{\theta^{2}}{\theta^{2} + \sigma^{2}}$
    \item $C_{2,\min} = \tfrac{\mu^{2}n + \theta^{2}}{\mu^{2}n + \theta^{2} + \sigma^{2}}, C_{3,\min} = \tfrac{\theta^{2}}{\theta^{2} + \sigma^{2}}$ \ .
  \end{enumerate}
  Since it is $\tfrac{\mu^{2}n + \theta^{2}}{\mu^{2}n + \theta^{2} + \sigma^{2}} \geq \tfrac{\theta^{2}}{\theta^{2} + \sigma^{2}}$ the case $C_{2,\min} = \tfrac{\mu^{2}n + \theta^{2}}{\mu^{2}n + \theta^{2} + \sigma^{2}} , C_{3,\min} = \rho_{N,\omega}$ does not occur. \\
  For the first case, see the analysis for $k<n$ and $C_{3,\min} = \rho_{N,\omega}$. \\
  For the second case, plugging in $\rho_{N,\omega}$, the risk is given by
  \begin{align}\label{unrolling_iid_even_second_case}
   &\tfrac{\mu^{2}}{2}n(1 - \omega)^{2N} + \tfrac{\theta^{2}}{2}(1 - \omega)^{2N} +  \tfrac{\theta^{2}}{2}(n-1)\tfrac{\sigma^{4}}{(\theta^{2} + \sigma^{2})^{2}} + \\
   &\tfrac{\sigma^{2}}{2}(1-(1-\omega)^{N})^{2} + \tfrac{\sigma^{2}}{2}(n-1)\tfrac{\theta^{4}}{(\theta^{2} + \sigma^{2})^{2}} \nonumber \ .
  \end{align}
  An optimization over $\omega \in \Big[1-\sqrt[N]{\tfrac{\sigma^{2}}{\theta^{2} + \sigma^{2}}},1-\sqrt[N]{\tfrac{\sigma^{2}}{\mu^{2}n + \theta^{2} + \sigma^{2}}}\Big[\ \bigcup\ \Big]1+\sqrt[N]{\tfrac{\sigma^{2}}{\mu^{2}n + \theta^{2} + \sigma^{2}}},1+\sqrt[N]{\tfrac{\sigma^{2}}{\theta^{2} + \sigma^{2}}}\Big]$ (which corresponds to $C_{2,\min} = \rho_{N,\omega}$ and $C_{3,\min} = \tfrac{\theta^{2}}{\theta^{2} + \sigma^{2}}$) shows that the infimum of (\ref{unrolling_iid_even_second_case}) is given by 
  \begin{align*}
    \tfrac12 \tfrac{\sigma^{2}\theta^{2}}{\sigma^{2} + \theta^{2}} + \tfrac12 \tfrac{\sigma^{2}(\mu^{2}n + \theta^{2})}{\mu^{2}n + \theta^{2} + \sigma^{2}}
  \end{align*}
  with $\arginf_{\omega} \cE_{i.i.d., k=n} = 1\pm \sqrt[N]{\tfrac{\sigma^{2}}{\mu^{2}n + \theta^{2} + \sigma^{2}}}$ which is not attained. 
  For the third case, the risk is independent of $\omega$ and after some transformations we get 
  \begin{align*}
    \tfrac12 \tfrac{\sigma^{2}\theta^{2}}{\sigma^{2} + \theta^{2}} + \tfrac12 \tfrac{\sigma^{2}(\mu^{2}n + \theta^{2})}{\mu^{2}n + \theta^{2} + \sigma^{2}}\ ,
  \end{align*}
  which is equal to the infimum of the second case. With $C_{2,\min} = \tfrac{\mu^{2}n + \theta^{2}}{\mu^{2}n + \theta^{2} + \sigma^{2}}$ and $C_{3,\min} = \tfrac{\theta^{2}}{\theta^{2} + \sigma^{2}}$ corresponding to $1-\sqrt[N]{\tfrac{\sigma^{2}}{\mu^{2}n + \theta^{2} + \sigma^{2}}}\leq\omega\leq1+\sqrt[N]{\tfrac{\sigma^{2}}{\mu^{2}n + \theta^{2} + \sigma^{2}}}$, it holds that the optimized risk is attained for $\omega \in \Big[1-\sqrt[N]{\tfrac{\sigma^{2}}{\mu^{2}n + \theta^{2} + \sigma^{2}}},1+\sqrt[N]{\tfrac{\sigma^{2}}{\mu^{2}n- + \theta^{2} + \sigma^{2}}}\Big]$. 
  \item $N$ odd: \\
  In this case, again, we analyze the optimized risk numerically. 
\end{itemize}

\section{Further proofs}\label{app:further-proofs}

\begin{proof}[Proof of Lemma~\ref{lem:best-est-constant}]
  It is 
  \begin{align*}
    \cE_{const.}(\T) &= \tfrac{\mu^{2}+ \theta^{2}}2\norm{(\T-\I)\1}_{2}^{2} + \tfrac{\sigma^{2}}2\norm{\T}_{F}^{2}\\ 
    &= \sum_{k=1}^{n} \tfrac{\mu^{2}+\theta^{2}}2(\scp{\ttt_{k}}{\1}-1)^{2} + \tfrac{\sigma^{2}}{2}\norm{\ttt_{k}}^{2}_{2},
  \end{align*}
  where $\ttt_{k}$ is the $k$-th row vector of $\T$.  The minimization separates over the rows and we get the optimality condition 
  \begin{align*}
    (\mu^{2}+\theta^{2})(\scp{\ttt_{k}}{\1} - 1)\1 + \sigma^{2} \ttt_{k} = 0,
  \end{align*}
  for every $k=1,\ldots,n$. By taking the inner product with $\1$ we get 
  \begin{align*}
    \scp{\ttt_{k}}{\1}=\tfrac{(\mu^{2}+\theta^{2})n}{\big((\mu^{2}+\theta^{2
    })n+\sigma^{2})}.
  \end{align*}
  Plugging this in the optimality condition, we get for each row $\ttt_{k}$
  \begin{align*}
    \ttt_{k} & = \tfrac{\mu^{2}+\theta^{2}}{\sigma^{2}}(1 -\tfrac{(\mu^{2}+\theta^{2})n}{(\mu^{2}+\theta^{2})n+\sigma^{2}})\1\\
    & = \tfrac{\mu^{2}+\theta^{2}}{n(\mu^{2}+\theta^{2}) + \sigma^{2}}\1.
  \end{align*}
  Thus, the best linear estimator is given by the matrix 
  \begin{align*}
    \T^{*} = \tfrac{\mu^{2}+\theta^{2}}{n(\mu^{2}+\theta^{2}) + \sigma^{2}}\1\1^{\top}.
  \end{align*}
  It is left to calculate the corresponding risk. We get 
  \begin{align*}
    \T^{*}\1 = \tfrac{n(\mu^{2}+\theta^{2})}{n(\mu^{2}+\theta^{2})+\sigma^{2}}\1
  \end{align*}
  and hence 
  \begin{align*}
    \norm{(\T^{*}-\I)\1}_{2}^{2} & = n\Big(\tfrac{n(\mu^{2}+\theta^{2})}{n(\mu^{2}+\theta^{2})+\sigma^{2}}-1\Big)^{2}\nonumber\\
    & = n\tfrac{\sigma^{4}}{(n(\mu^{2}+\theta^{2})+\sigma^{2})^{2}}.
  \end{align*}
  Moreover we have 
  \begin{align*}
    \norm{\T^{*}}_{F}^{2} = n^{2}\tfrac{(\mu^{2}+\theta^{2})^{2}}{\big(n(\mu^{2}+\theta^{2})+\sigma^{2}\big)^{2}}
  \end{align*}
  which gives by Corollary~\ref{cor:true-risks}
  \begin{align*}
    \min_{\T}\E_{\substack{\bm \epsilon\sim\cN\\\y\sim\cD}} \tfrac12\norm{\T(\y+\bm \epsilon)-\y}_{2}^{2} & =\tfrac{\mu^{2}+\theta^{2}}{2} n\tfrac{\sigma^{4}}{(n(\mu^{2}+\theta^{2})+\sigma^{2})^{2}}  + \tfrac{\sigma^{2}}{2}n^{2}\tfrac{(\mu^{2}+\theta^{2})^{2}}{\big(n(\mu^{2}+\theta^{2})+\sigma^{2}\big)^{2}}\\
    & =\tfrac{n\sigma^{2}(\mu^{2}+\theta^{2})}{2}\Big[\tfrac{\sigma^{2}}{(n(\mu^{2}+\theta^{2})+\sigma^{2})^{2}} + \tfrac{n(\mu^{2}+\theta^{2})}{\big(n(\mu^{2}+\theta^{2}) + \sigma^{2}\big)^{2}}\Big]\\
    & =              \tfrac{\sigma^{2}}{2}\tfrac{n(\mu^{2}+\theta^{2})}{n(\mu^{2}+\theta^{2})+\sigma^{2}}                                                                                                
  \end{align*}
  as claimed.
\end{proof}

\begin{proof}[Proof of Lemma~\ref{lem:best-est-iid}]
  We start by calculating the gradient of $\cE_{i.i.d}$ with respect to $\T$ and setting it to zero.
  We get
\begin{align*}
  \bm 0 & = \mu^{2}(\T-\I)\1\1^{\top} + \theta^{2}(\T-\I) + \sigma^{2}\T\\
  & = \T(\mu^{2}\1\1^{\top} + (\theta^{2} +\sigma^{2})\I) - \mu^{2}\1\1^{\top} - \theta^{2} \I,
\end{align*}
which is equivalent to 
\begin{align*}
\T = ((\theta^{2} +\sigma^{2})\I + \mu^{2}\1\1^{\top})^{-1} (\theta^{2} \I + \mu^{2}\1\1^{\top}),
\end{align*}
if $(\theta^{2} +\sigma^{2})\I + \mu^{2}\1\1^{\top}$ is invertible.

Let $a,b$ and $c$ be real numbers. If $a\I + b\1\1^{\top}$ is invertible, i.e. $a+nb\neq 0$ and $a\neq 0$, it is $(a\I + b\1\1^{\top})^{-1} = \tfrac1a \I - \tfrac{b}{a(a+nb)}\1\1^{\top}$. Moreover, it holds that 
\begin{align*}
  (a\I+b\1\1^{\top})^{-1}((a-c)\I + b\1\1^{\top}) & = \Big(\tfrac1a \I - \tfrac{b}{a(a+nb)}\1\1^{\top}\Big)((a-c)\I+b\1\1^{\top})\\
                                          & = \tfrac{a-c}{a} \I + \Big(\tfrac{b}{a} - \tfrac{b(a-c)}{a(a+nb)} - \tfrac{nb^{2}}{a(a+nb)}\Big)\1\1^{\top}\\
                                          & = \tfrac1a\Big((a-c)\I + \tfrac{b(a+nb) - b(a-c) - nb^{2}}{a+nb}\1\1^{\top}\Big)\\
                                          & = \tfrac1a\Big((a-c)\I + \tfrac{bc}{a+nb}\1\1^{\top}\Big).
\end{align*}
With $a = \theta^{2} + \sigma^{2}$, $b = \mu^{2}$ and $c = \sigma^{2}$ we get \begin{align*}
  \T^{*} & = \tfrac{1}{\theta^{2}+\sigma^{2}}\Big(\theta^{2} \I + \tfrac{\sigma^{2}\mu^{2}}{\theta^{2}+\sigma^{2}+n\mu^{2}}\1\1^{\top}\Big)\\
  & =\tfrac{\theta^{2}}{\theta^{2}+\sigma^{2}} \I +  \tfrac{\sigma^{2}}{\theta^{2}+\sigma^{2}}\tfrac{\mu^{2}}{n\mu^{2} + \theta^{2}+\sigma^{2}}\1\1^{\top}.
\end{align*}
After we calculated the best estimator it is left to calculate the corresponding risk. Therefore, we use 
\begin{align}\label{eq:help-best-linear-iid}
  \T^{*} - \I = \tfrac{\sigma^{2}}{\theta^{2}+\sigma^{2}}\Big(\tfrac{\mu^{2}}{n\mu^{2} +\theta^{2}+\sigma^{2}}\1\1^{\top} -\I\Big)
\end{align}
to calculate the three norms in $\cE(\T^{*})$. We have 
\begin{align*}
  \norm{(\T^{*}-\I)\1}^{2} & = n \tfrac{\sigma^{4}}{(\theta^{2} + \sigma^{2})^{2}}\Big (\tfrac{n\mu^{2}}{n\mu^{2} + \theta^{2} + \sigma^{2}} - 1 \Big)^{2}\\
   & = \tfrac{n\sigma^{4}}{(n\mu^{2} + \theta^{2}+\sigma^{2})^2}.
\end{align*}
and 
\begin{align*}
  \norm{\T^{*}-\I}_{F}^{2} & = \tfrac{\sigma^{4}}{(\theta^{2} + \sigma^{2})^{2}}\Big(n(\tfrac{\mu^{2}}{n\mu^{2} + \theta^{2} + \sigma^{2}} - 1)^{2} + n(n - 1)(\tfrac{\mu^{2}}{n\mu^{2} +\theta^{2} + \sigma^{2}})^{2}\Big)\\
  & = \sigma^{4}\Big(\tfrac{1}{(n\mu^{2} + \theta^{2}+\sigma^{2})^2} + \tfrac{(n-1)}{(\theta^{2} + \sigma^{2})^2}\Big)
\end{align*}
by using~(\ref{eq:help-best-linear-iid}). Furthermore, it is by Corollary~\ref{cor:true-risks}
\begin{align*}
  \norm{\T^{*}}_{F}^{2} & = n\Big(\tfrac{\theta^{2}}{\theta^{2} + \sigma^{2}} + \tfrac{\sigma^{2}}{\theta^{2} + \sigma^{2}}\tfrac{\mu^{2}}{n\mu^{2} + \theta^{2} + \sigma^{2}}\Big)^{2} + n(n-1)\Big(\tfrac{\sigma^{2}}{\theta^{2} + \sigma^{2}}\tfrac{\mu^{2}}{n\mu^{2} +\theta^{2} + \sigma^{2}}\Big)^2\\
  & = (n-1)\tfrac{\theta^{4}}{(\theta^{2}+\sigma^{2})^{2}} + \Big(\tfrac{\theta^{2}}{\theta^{2}+\sigma^{2}} + \tfrac{\sigma^{2}}{\theta^{2}+\sigma^{2}}\tfrac{n\mu^{2}}{n\mu^{2} + \theta^{2}+\sigma^{2}}\Big)^{2}.
\end{align*}
Plugging this in the formular for the risk of the best linear estimator $\T^{*}$ we get
\begin{align*}
    \cE_{i.i.d.}(\T^{*}) & = \tfrac{\mu^{2}}2\tfrac{n\sigma^{4}}{(n\mu^{2} + \theta^{2}+\sigma^{2})^2} + \tfrac{\theta^{2}}2\sigma^{4}\Big(\tfrac{1}{(n\mu^{2} + \theta^{2}+\sigma^{2})^2} + \tfrac{(n-1)}{(\theta^{2} + \sigma^{2})^2}\Big)\\ 
    & \quad +  \tfrac{\sigma^{2}}2\Big((n-1)\tfrac{\theta^{4}}{(\theta^{2}+\sigma^{2})^{2}} + \Big(\tfrac{\theta^{2}}{\theta^{2}+\sigma^{2}} + \tfrac{\sigma^{2}}{\theta^{2} + \sigma^{2}}\tfrac{n\mu^{2}}{n\mu^{2} + \theta^{2} + \sigma^{2}}\Big)^{2}\Big)\\
  & = \tfrac{\sigma^{2}}2\Big[\tfrac{n\mu^{2}\sigma^{2}}{(n\mu^{2} + \theta^{2} + \sigma^{2})^{2}}\Big(1 + \tfrac{n\mu^{2}\sigma^{2}}{(\theta^{2} + \sigma^{2})^{2}}\Big)\\ 
  & \quad + \tfrac{\theta^{2}}{(\theta^{2} + \sigma^{2})^{2}}\Big(\tfrac{n\mu^{2}\sigma^{2}}{n\mu^{2} + \theta^{2} + \sigma^{2}} + n\theta^{2} + \sigma^{2}\Big((n-1) + \tfrac{(\theta^{2} + \sigma^{2})}{(n\mu^{2} + \theta^{2} + \sigma^{2})^{2}}\Big)\Big)\Big]\\
  & = \tfrac{\sigma^{4}\theta^{2}n}{2(\theta^{2}+\sigma^{2})^{2}}\Big(\tfrac{\theta^{2}}{\sigma^{2}} + 1 -\tfrac1n \\
  & \quad + \tfrac{\mu^{2}}{n\mu^{2}+\theta^{2}+\sigma^{2}}\left(\tfrac1{n\mu^{2}+\theta^{2}+\sigma^{2}}\left(\tfrac{n\sigma^{2}}{\theta^{2}} + (\theta^{2} + \sigma^{2})^{2}(\tfrac1{\theta^{2}} + \tfrac1{n\mu^{2}})\right)+ 2\right)\Big).
\end{align*}
\end{proof}

\begin{proof}[Proof of Theorem~\ref{thm:best-bilevel-constant}]
  Using singular value decomposition we can write 
  \begin{align*}
    \RRR^{\top}\RRR = \V\Sigmaa \V^{\top}
  \end{align*}
  with orthonormal $\V\in \R^{n\times n}$ and $\Sigmaa = \diag(\sigma_{1},\ldots,\sigma_{n})$ fulfilling $\sigma_{1}\geq \sigma_{2}\geq\cdots \geq\sigma_{k}\geq \sigma_{k+1}= \cdots=\sigma_{n}=0$. Plugging this into the objective we get 
  \begin{align*}
    \cE_{const.}(\T) &=  \tfrac{\mu^{2}+\theta^{2}}{2}\sum_{j=1}^{k}\tfrac{\sigma_{j}^{2}}{(1+\sigma_{j})^{2}}\abs{\scp{\vv_{j}}{\1}}^{2} + \tfrac{\sigma^{2}}{2}\Big(\sum_{j=1}^{k}\tfrac{1}{(1+\sigma_{j})^{2}} + n-k\Big)\\
    & = \tfrac12\sum_{j=1}^{k}\tfrac{(\mu^{2}+\theta^{2})\abs{\scp{\vv_{j}}{\1}}^{2}\sigma_{j}^{2} + \sigma^{2}}{(1+\sigma_{j})^{2}} + \tfrac{\sigma^{2}}{2}(n-k).
  \end{align*}
  Since the first sum is positive, the value $\sigma^{2}(n-k)/2$ is an
  obvious lower bound. By choosing $\vv_{j}$ such that
  $\scp{\vv_{j}}{\1}=0$ for $j=1,\dots,k$ (which is possible if $k<n$)
  and letting $\sigma_{j}\to \infty$ we see that the first sum can be
  arbitrarily small. 
  
  In the case $k=n$ we have
  \begin{align*}
    \cE_{const.}(\T) &= \tfrac12\sum_{j=1}^{n}\tfrac{(\mu^{2}+\theta^{2})\abs{\scp{\vv_{j}}{\1}}^{2}\sigma_{j}^{2} + \sigma^{2}}{(1+\sigma_{j})^{2}}\\ 
    &= \tfrac12\sum_{j=1}^{n}\tfrac{(\mu^{2}+\theta^{2})\sigma_{j}^{2}}{(1+\sigma_{j})^{2}}\abs{\scp{\vv_{j}}{\1}}^{2} + \tfrac12\sum_{j=1}^{n}\tfrac{\sigma^{2}}{(1+\sigma_{j})^{2}} \ .
  \end{align*}
  Applying Lemma~\ref{lem:unrolling_orthonormal} with $c_{j} = \tfrac{(\mu^{2} + \theta^{2})\sigma_{j}^{2}}{(1+\sigma_{j})^{2}}\geq 0$, we have that $\vv_{j^{*}} = \tfrac1{\sqrt{n}}\1$ with $\abs{\scp{\vv_{j^{*}}}{\1}}^{2} = n$ and $\vv_{j} \perp \1$ for $j\neq j^{*}$ (at this point we do not know the value of $j^{*}$, but this does not matter for the following). Thus, the right side is reduced to 
  \begin{align*}
    \tfrac12\tfrac{(\mu^{2}+\theta^{2})n\sigma_{j^{*}}^{2} + \sigma^{2}}{(1+\sigma_{j^{*}})^{2}} + \tfrac12\sum_{j\neq j^{*}}^{n}\tfrac{\sigma^{2}}{(1+\sigma_{j})^{2}}\ .
  \end{align*}
  The second term has infimum equal to $0$ which is approximated (and not reached) for $\sigma_{j}\to\infty$. Thus, we have 
  \begin{align*}
     \cE_{const.}(\T) > 
     \tfrac12\tfrac{(\mu^{2}+\theta^{2})n\sigma_{j^{*}}^{2} + \sigma^{2}}{(1+\sigma_{j^{*}})^{2}} \ .
  \end{align*}
  Minimizing over $\sigma_{j^{*}}>0$ and using that the minimum of $(as^{2}+b)/(1+s)^{2}$ is $ab/(a+b)$ we get the sharp lower bound
  \begin{align*}
    \cE_{const.}(\T) > \tfrac{1}{2}\tfrac{n\sigma^{2}}{n + \tfrac{\sigma^{2}}{\mu^{2}+\theta^{2}}} = \tfrac{\sigma^{2}}2\tfrac{n(\mu^{2}+\theta)}{n(\mu^{2}+\theta^{2}) + \sigma^{2}}.
  \end{align*}

\end{proof}

\begin{proof}[Proof of Proposition~\ref{prop:best-bilevel-iid_estimators}]
  Consider the case with $k<n$ first. 
  Again we argue by using the representation
  \begin{align*}
    \T = (\I+\RRR^{\top}\RRR)^{-1} = \V
    \begin{bmatrix}
      \tfrac{1}{1+\sigma_{1}}\\
      & \ddots \\
      && \tfrac1{1+\sigma_{k}}\\
      &&& 1 \\
      &&&& \ddots\\
      &&&&& 1
    \end{bmatrix}\V^{\top}
  \end{align*}
  with orthonormal $\V$ from the proof of Theorem~\ref{thm:expressivity-bilevel} and plug $\T$ into $\cE_{i.i.d.}$ defined in Corollary~\ref{cor:true-risks} and get.
  \begin{align*}
    \cE_{i.i.d.}(\T) & = \tfrac{\mu^{2}}{2}\sum_{j=1}^{k}\tfrac{\sigma_{j}^{2}}{(1+\sigma_{j})^{2}}\abs{\scp{\vv_{j}}{\1}}^{2} + \tfrac{\theta^{2}}{2}\sum_{j=1}^{k} \tfrac{\sigma_{j}^{2}}{(1+\sigma_{j})^{2}} +  \tfrac{\sigma^{2}}{2}\Big(\sum_{j=1}^{k}\tfrac{1}{(1+\sigma_{j})^{2}} + n-k\Big)\\
    & = \tfrac12\sum_{j=1}^{k}\tfrac{(\mu^{2}\abs{\scp{\vv_{j}}{\1}}^{2}+\theta^{2})\sigma_{j}^{2} + \sigma^{2}}{(1+\sigma_{j})^{2}} + \tfrac{\sigma^{2}}{2}(n-k).
  \end{align*}
  To minimize the first sum, we first choose $\vv_{1},\dots,\vv_{k}\bot\1$
  such that all the inner products vanish and then we are left with
  minimizing the function
  $f(s) = \tfrac{\theta^{2} s^{2} + \sigma^{2}}{(1+s)^2}$ over $s$. This
  function has a unique minimum at $s = \tfrac{\sigma^{2}}{\theta^{2}}$
  and minimal value
  $\tfrac{\sigma^{2}\theta^{2}}{\theta^{2}+\sigma^{2}}$. Hence, we get as minimal value for the risk:
  \begin{align*}
    \min_{\T\in \cA_{k}} \cE_{i.i.d.}(\T)& = \tfrac12 k\tfrac{\sigma^{2}\theta^{2}}{\theta^{2}+\sigma^{2}} + \tfrac{\sigma^{2}}{2}(n-k)\\
    & = \tfrac{\sigma^{2}}{2}\Big(k\tfrac{\theta^{2}}{\theta^{2}+\sigma^{2}} + (n-k)\Big).
  \end{align*}
  It is left to show the result for $k=n$.  We now have 
   \begin{align*}
    \T = (\I+\RRR^{\top}\RRR)^{-1} = \V
    \begin{bmatrix}
      \tfrac{1}{1+\sigma_{1}}\\
      & \ddots \\
      && \tfrac1{1+\sigma_{n}}
    \end{bmatrix}\V^{\top}
  \end{align*}
with orthonormal $\V$ and get 
 \begin{align*}
    \cE_{i.i.d.}(\T) & = \tfrac{\mu^{2}}{2}\sum_{j=1}^{n}\tfrac{\sigma_{j}^{2}}{(1+\sigma_{j})^{2}}\abs{\scp{\vv_{j}}{\1}}^{2} + \tfrac{\theta^{2}}{2}\sum_{j=1}^{n} \tfrac{\sigma_{j}^{2}}{(1+\sigma_{j})^{2}} +  \tfrac{\sigma^{2}}{2}\sum_{j=1}^{n}\tfrac{1}{(1+\sigma_{j})^{2}} \\
    &= \tfrac{1}{2}\sum_{j=1}^{n}\tfrac{\mu^{2}\sigma_{j}^{2}}{(1+\sigma_{j})^{2}}\abs{\scp{\vv_{j}}{\1}}^{2} + \tfrac12\sum_{j=1}^{n}\tfrac{\theta^{2}\sigma_{j}^{2} + \sigma^{2}}{(1+\sigma_{j})^{2}} \ .
 \end{align*}
  Applying Lemma~\ref{lem:unrolling_orthonormal} with $c_{j} = \tfrac{\mu^{2}\sigma_{j}^{2}}{(1+\sigma_{j})^{2}}\geq 0$, we have that $\vv_{j^{*}} = \tfrac1{\sqrt{n}}\1$ with $\abs{\scp{\vv_{j^{*}}}{\1}}^{2} = n$ and $\vv_{j} \perp \1$ for $j\neq j^{*}$. Thus, the right side is reduced to 
  \begin{align*}
    \tfrac12\tfrac{\mu^{2}n\sigma_{j^{*}}^{2}+\theta^{2}\sigma_{j^{*}}^{2} + \sigma^{2}}{(1+\sigma_{j^{*}})^{2}} + \tfrac12\sum_{j\neq j^{*}}\tfrac{\theta^{2}\sigma_{1}^{2}+\sigma^{2}}{(1+\sigma_{j})^{2}}\ .
  \end{align*}
  An optimization over $\sigma_{j}>0$ for $j = 1,\ldots,n$, using that the minimum of $(as^{2}+b)/(1+s)^{2}$ is $ab/(a+b)$ and is attained at $b/a$, gives 
  \begin{align*}
    \cE_{const.}(\T) = \tfrac{\sigma^{2}}{2}\left((n-1)\tfrac{\theta^{2}}{\theta^{2} + \sigma^{2}} + \tfrac{n\mu^{2} + \theta^{2}}{n\mu^{2} + \theta^{2} + \sigma^{2}}\right).
  \end{align*}
  Moreover, the optimal values for $\sigma_{j}$ are
  \begin{align*}
    \sigma_{j} =
    \begin{cases}
      \tfrac{\sigma^{2}}{n\mu^{2} + \theta^{2}}, & j= j^{*}\\
     \tfrac{\sigma^{2}}{\theta^{2}} , & j\neq j^{*}.
    \end{cases}
  \end{align*}
  To conclude the proof we assume $j^{*} = n$  without loss of generality.
\end{proof}

\begin{proof}[Proof of Theorem~\ref{thm:best-unrolling-constant}]
  We start the proof with the situation of $k<n$. Again we use the SVD of $\RRR$ (more precisely, the representation $\RRR^{T}\RRR = \V\diag(\sigmaa)\V$), the substitution $\rho_{j} = f(\sigma_{j})$ and the abbreviation $\rho_{N,\omega} = 1-(1-\omega)^{N}$ and get that
  \begin{align*}
    \T = \V
  \begin{bmatrix}
  \rho_{1}\\
    & \ddots\\
    && \rho_{k}\\
    &&& \rho_{N,\omega}\\
    &&&&\ddots\\
    &&&&& \rho_{N,\omega}
  \end{bmatrix}
  \V^{\top}.
  \end{align*}
  Plugging this in the objective we get
  \begin{align*}
    \cE_{const.}(\T) & =\tfrac{\mu^{2} +\theta^{2}}{2}\left(\sum_{j=1}^{k}\big(\rho_{j}-1\big)^{2}\abs{\scp{\vv_{j}}{\1}}^{2} + \sum_{j=k+1}^{n}(\rho_{N,\omega}-1)^{2}\abs{\scp{\vv_{j}}{\1}}^{2}\right)\\
           & \quad + \tfrac{\sigma^{2}}{2}\left(\sum_{j=1}^{k}\rho_{j}^{2} + (n-k)\rho_{N,\omega}^{2}\right)
  \end{align*}
  Applying Lemma~\ref{lem:unrolling_orthonormal} with $c_{j} = (\rho_{j}-1)^{2}$ for $j = 1,\ldots, k$ and $(\rho_{N,\omega}-1)^{2}$ for $j = k+1, \ldots, n$, an optimal choice of $\V$ corresponds to $\abs{\scp{\vv_{j^{*}}}{\1}}^{2} = n$ and $\abs{\scp{\vv_{j}}{\1}}^{2} = 0, j \neq j^{*}$ for some $j^{*}\in \{1,\ldots,n\}$. Since $\abs{\scp{\vv_{j}}{\1}}^{2}$ appears in two different sums, we have to distinguish two cases. The first case $j^{*}\in \{1,\ldots,k\}$ occurs if $(\rho_{j^{*}}-1)^{2}\leq (\rho_{N,\omega}-1)^{2}$ while the second case $j^{*}\in \{k+1, \ldots,n\}$ occurs if $(\rho_{j}-1)^{2}\geq (\rho_{N,\omega}-1)^{2}$ for all $j\in \{1,\ldots,k\}$. We start with the case distinction in the situation of $N$ even. 
 \begin{description}
   \item[$N$ even:]  From Theorem~\ref{thm:expressivity-unrolling} we know that for $N$ even we have $0\leq\rho_{j}\leq 1-(1-\omega)^{N} = \rho_{N,\omega}$.  
   \begin{description}
   \item[$j^{*} \in \{1,\ldots,k\}$:] In this case we have
     \begin{align}\label{eq:unrolling_const_j_1}
       \cE_{const.}(\T) & =\tfrac{n(\mu^{2} +\theta^{2})}{2}(\rho_{j^{*}}-1)^{2} + \tfrac{\sigma^{2}}{2}\Big(\sum_{j=1}^{k}\rho_{j}^{2} + (n-k)\rho_{N,\omega}^{2}\Big).
     \end{align} 
     An optimization over $\rho_{j}\leq \rho_{N,\omega}$ gives the optimal values $\rho_{j^{*}}^{*} = \min\left(\tfrac{n(\mu^{2}+\theta^{2})}{n(\mu^{2}+\theta^{2}) + \sigma^{2}}, \rho_{N,\omega}\right)$ and $\rho_{j}^{*} = 0$ for all $j \in \{1,\ldots,k\}\setminus{\{j^{*}\}}$. Suppose that the minimum in $\rho_{j^{*}}^{*}$ is attained at $\tfrac{n(\mu^{2}+\theta^{2})}{n(\mu^{2}+\theta^{2}) + \sigma^{2}}$. Since both terms, $\tfrac{n(\mu^{2}+\theta^{2})}{n(\mu^{2}+\theta^{2}) + \sigma^{2}}$ as well as $\rho_{N,\omega}$, are in $[0,1]$ we have 
     \begin{align*}
       \left(\tfrac{n(\mu^{2}+\theta^{2})}{n(\mu^{2}+\theta^{2}) + \sigma^{2}}-1\right)^{2}=(\rho_{j^{*}}^{*}-1)^{2}>(\rho_{N,\omega}-1)^{2}\ .
     \end{align*}
     This is a contradiction to $j^{*}\in\{1,\ldots,k\}$, since in this case we have $(\rho_{j^{*}}-1)^{2}\leq(\rho_{N,\omega}-1)^{2}$.
     Thus, we get that $\rho_{j^{*}}=\rho_{N,\omega}$ with corresponding risk 
     \begin{align}\label{eq:unrolling_const_j_1_min_even}
       \cE_{i.i.d.}(\T) = \tfrac{\mu^{2} +\theta^{2}}{2}(\rho_{N,\omega}-1)^{2}n + \tfrac{\sigma^{2}}{2}\Big(\rho_{N,\omega}^{2} + (n-k)\rho_{N,\omega}^{2}\Big).
     \end{align}
   \item[$j^{*} \in \{k+1,\ldots,n\}$:] In that case, it is 
     \begin{align}\label{eq:unrolling-const-j-n}
       \cE_{const.}(\T) & =\tfrac{\mu^{2} +\theta^{2}}{2}(\rho_{N,\omega}-1)^{2}n + \tfrac{\sigma^{2}}{2}\left(\sum_{j=1}^{k}\rho_{j}^{2} + (n-k)\rho_{N,\omega}^{2}\right).
     \end{align}
     Since $\rho_{j}$ only appears as $\rho_{j}^{2}$, minimizing over $\rho_{j}$ gives $\rho_{j}^{*} = 0$ for all $j\in\{1,\ldots,k\}$ fulfilling the bounds from Theorem~\ref{thm:expressivity-unrolling} and the corresponding risk is given by 
     \begin{align}\label{eq:unrolling-const-j-n-min-even}
       \cE_{const.}(\T)=\tfrac{\mu^{2} + \theta^{2}}{2}n(\rho_{N,\omega}-1)^{2} + \tfrac{\sigma^{2}}{2}(n-k)\rho_{N,\omega}^{2} \ .
     \end{align}
   \end{description}
   Concluding the case distinction in $j^{*}$, we observe that the expression for $\cE_{const.}(\T)$ in  (\ref{eq:unrolling-const-j-n-min-even}) is smaller than the one in (\ref{eq:unrolling_const_j_1_min_even}). Hence, the case $j^{*}\in\{k+1,\ldots,n\}$ is always better than $j^{*}\in \left\{ 1,\ldots,k \right\}$, which proves the claim for $N$ even and $k<n$.
 \item[$N$ odd:] For $N$ odd we have from Theorem~\ref{thm:expressivity-unrolling} that $\rho_{j} \geq c_{N,\omega}$. 
  \begin{description}
    \item[$j^{*} \in \{1,\ldots,k\}$:] In this case we also have to minimize $\cE_{const.}$ from (\ref{eq:unrolling_const_j_1}) and can compute that a minimization over $\rho_{j}\geq c_{N,\omega}$ gives $\rho_{j^{*}}^{*} = \max\left\{\tfrac{n(\mu^{2}+\theta^{2})}{n(\mu^{2}+\theta^{2}) + \sigma^{2}}, c_{N,\omega}\right\}$ and $\rho_{j}^{*} = c_{N,\omega}$ for all $j \in \{1,\ldots, k\}\setminus{\{j^{*}\}}$. The corresponding risk is 
    \begin{align}\label{eq:unrolling_const_j_1_min_odd}
      \cE_{const.}(\T) &= \tfrac{n(\mu^{2} +\theta^{2})}{2}\left(\max\{\tfrac{n(\mu^{2}+\theta^{2})}{n(\mu^{2}+\theta^{2}) + \sigma^{2}}, c_{N,\omega}\}-1\right)^{2}\\
       &\quad + \tfrac{\sigma^{2}}{2}\left(\max\{\tfrac{n(\mu^{2}+\theta^{2})}{n(\mu^{2}+\theta^{2}) + \sigma^{2}}, c_{N,\omega}\}^{2} + (k-1)c_{N,\omega}^{2} + (n-k)\rho_{N,\omega}^{2}\right)\ . \nonumber
    \end{align}
    Note that this case only occurs, if $(\rho_{j}^{*}-1)^{2}\leq (\rho_{N,\omega}-1)^{2}$.
    \item[$j^{*}\in\{k+1,\ldots, n\}$:] Again,  an optimization of (\ref{eq:unrolling-const-j-n}) over $\rho_{j}\geq c_{N,\omega}$ gives $\rho_{j}^{*} = c_{N,\omega}$ for all $j \in \{1,\ldots,k\}$ and thus 
    \begin{align}\label{eq:unrolling_const_j_n_min_odd}
      \cE_{const.}(\T) = \tfrac{\mu^{2} +\theta^{2}}{2}(\rho_{N,\omega}-1)^{2}n + \tfrac{\sigma^{2}}{2}\left(kc_{N,\omega}^{2} + (n-k)\rho_{N,\omega}^{2}\right)\ .
    \end{align}
    This only occurs if $(c_{N,\omega}-1)^{2}\geq (\rho_{N,\omega}-1)^{2}$.
  \end{description}
  Thus, for $N$ odd and $k<n$ the optimal risk is obtained for the minimum of (\ref{eq:unrolling_const_j_1_min_odd}) and (\ref{eq:unrolling_const_j_n_min_odd}), i.e. we have after some reformulation
  \begin{alignat*}{2}
    \cE_{const.}(\T) &= \tfrac{\sigma^{2}}{2}\Big((k-&&1)c_{N,\omega}^{2} + (n-k)\rho_{N,\omega}^{2}\Big)\\
     &\quad+ \min\Big\{&&\tfrac{\mu^{2} +\theta^{2}}{2}\Big(\max\left\{\tfrac{n(\mu^{2}+\theta^{2})}{n(\mu^{2}+\theta^{2}) + \sigma^{2}}, c_{N,\omega}\right\}-1\Big)^{2}n + \tfrac{\sigma^{2}}{2}\max\left\{\tfrac{n(\mu^{2}+\theta^{2})}{n(\mu^{2}+\theta^{2}) + \sigma^{2}},  c_{N,\omega}\right\}^{2}, \\
     &\phantom{=}&&\tfrac{n(\mu^{2} +\theta^{2})}{2}(\rho_{N,\omega}-1)^{2} + \tfrac{\sigma^{2}}{2}c_{N,\omega}^{2}\Big\}
  \end{alignat*}
  which proves the claim for $N$ odd and $k<n$.
  \end{description}
  It is left to show the result for $k=n$. In this case the risk is given by
  \begin{align*}
  \cE_{const.}(\T) & =\tfrac{\mu^{2} +\theta^{2}}{2}\sum_{j=1}^{n}\big(\rho_{j}-1\big)^{2}\abs{\scp{\vv_{j}}{\1}}^{2} + \tfrac{\sigma^{2}}{2}\sum_{j=1}^{n}\rho_{j}^{2}
  \end{align*} 
  and no case distinction in $j^{*}$ is needed. Applying Lemma~\ref{lem:unrolling_orthonormal} with $c_{j} = (\rho_{j}-1)^{2}$ we get 
  \begin{align*}
    \cE_{const.}(\T) = \tfrac{n(\mu^{2} +\theta^{2})}{2}(\rho_{j^{*}}-1\big)^{2}+ \tfrac{\sigma^{2}}{2}\sum_{j=1}^{n}\rho_{j}^{2} \ .
  \end{align*}
  For $N$ even we optimize over $\rho_{j}\leq \rho_{N,\omega}$ and get $\rho_{j^{*}}^{*} = \min\left\{\tfrac{n(\mu^{2}+\theta^{2})}{n(\mu^{2}+\theta^{2})+\sigma^{2}},\rho_{N,\omega}\right\}$ and $\rho_{j}^{*} = 0$ for all $j\neq j^{*}$. Thus, the optimal risk for $N$ even is 
\begin{align*}
  \cE_{const.}(\T) = \tfrac{n(\mu^{2} +\theta^{2})}{2}(\min\left\{\tfrac{n(\mu^{2}+\theta^{2})}{n(\mu^{2}+\theta^{2})+\sigma^{2}},\rho_{N,\omega}\right\}-1)^{2}\ + \tfrac{\sigma^{2}}{2}\min\left\{\tfrac{n(\mu^{2}+\theta^{2})}{n(\mu^{2}+\theta^{2})+\sigma^{2}},\rho_{N,\omega}\right\}^{2}.
\end{align*}
For $N$ odd, an optimization over $\rho_{j} \geq c_{N,\omega}$ results in $\rho_{j^{*}}^{*} = \max\left\{\tfrac{n(\mu^{2}+\theta^{2})}{n(\mu^{2}+\theta^{2})+\sigma^{2}},c_{N,\omega}\right\}$, $\rho_{j}^{*} = c_{N,\omega}$ for $j\neq j^{*}$ and 
\begin{align*}
    \cE_{const.}(\T) &= \tfrac{n(\mu^{2} +\theta^{2})}{2}\left(\max\left\{\tfrac{n(\mu^{2}+\theta^{2})}{n(\mu^{2}+\theta^{2})+\sigma^{2}},c_{N,\omega}\right\}-1\right)^{2}\ \\
    &\quad + \tfrac{\sigma^{2}}{2}\max\left\{\tfrac{n(\mu^{2}+\theta^{2})}{n(\mu^{2}+\theta^{2})+\sigma^{2}},c_{N,\omega}\right\}^{2} + \tfrac{\sigma^{2}}{2}(n-1)c_{N,\omega}^{2}
\end{align*}
which proves the remaining claim for $k=n$.
\end{proof}

\begin{proof}[Proof of Theorem~\ref{thm:best-unrolling-iid}]
  We proceed analogously to the proof of Theorem~\ref{thm:best-unrolling-constant}. Thus, we consider $k<n$ first. For $\T$ of the form~(\ref{eq:estimator-unrolling}) we have again the representation 
  \begin{align*}
    \T = \V
  \begin{bmatrix}
  \rho_{1}\\
    & \ddots\\
    && \rho_{k}\\
    &&& \rho_{N,\omega}\\
    &&&&\ddots\\
    &&&&& \rho_{N,\omega}
  \end{bmatrix}
  \V^{\top},
  \end{align*}
  using~(\ref{eq:unrolling_estimator_svd}) and substituting $\rho_{j} = f(\sigma_{j})$. Plugging this in the objective function for the situation of random i.i.d. vectors from Corollary~\ref{cor:true-risks} we get 
  \begin{align*}
   \cE_{i.i.d.}(\T) & =\tfrac{\mu^{2}}{2}\left(\sum_{j=1}^{k}\big(\rho_{j}-1\big)^{2}\abs{\scp{\vv_{j}}{\1}}^{2} + \sum_{j=k+1}^{n}(\rho_{N,\omega}-1)^{2}\abs{\scp{\vv_{j}}{\1}}^{2}\right)\\
           & \quad + \tfrac{\theta^{2}}{2}\left(\sum_{j=1}^{k}\big(\rho_{j}-1\big)^{2} + \sum_{j=k+1}^{n}(\rho_{N,\omega}-1)^{2}\right) + \tfrac{\sigma^{2}}{2}\left(\sum_{j=1}^{k}\rho_{j}^{2} + (n-k)\rho_{N,\omega}^{2}\right).  
  \end{align*}
  Applying Lemma~\ref{lem:unrolling_orthonormal} with $c_{j} = (\rho_{j}-1)^{2}$ for $j = 1,\ldots, k$ and $(\rho_{N,\omega}-1)^{2}$ for $j = k+1, \ldots, n$ like in the proof of Theorem~\ref{thm:best-unrolling-constant}, an optimal choice of $\V$ again corresponds to $\abs{\scp{\vv_{j^{*}}}{\1}}^{2} = n$ and $\abs{\scp{\vv_{j}}{\1}}^{2} = 0, j \neq j^{*}$ for some $j^{*}\in \{1,\ldots,n\}$. Moreover, the same two cases have to be distinguished, namely $j^{*}\in \{1,\ldots,k\}$ and $j^{*}\in \{k+1, \ldots,n\}$. We start with the situation of $N$ even. 
 \begin{description}
   \item[$N$ even:]  From Theorem~\ref{thm:expressivity-unrolling} we already have that for $N$ even it is $0\leq\rho_{j}\leq \rho_{N,\omega}$.  
   \begin{description}
     \item[$j^{*} \in\{1,\ldots,k\}$:] In this case we have
     \begin{align}\label{eq:unrolling_iid_j_1}
    \cE_{i.i.d.}(\T) & =\tfrac{\mu^{2}}{2}n(\rho_{j^{*}}-1)^{2} + \tfrac{\theta^{2}}{2}\left(\sum_{j=1}^{k}(\rho_{j}-1)^{2} + (n-k)(\rho_{N,\omega}-1)^{2}\right)\\
     &\quad +  \tfrac{\sigma^{2}}{2}\left(\sum_{j=1}^{k}\rho_{j}^{2} + (n-k)\rho_{N,\omega}^{2}\right). \nonumber
  \end{align} 
  Optimizing over $\rho_{j}\leq \rho_{N,\omega}$ gives $\rho_{j}^{*} = \min\left\{\tfrac{\theta^{2}}{\theta^{2} + \sigma^{2}}, \rho_{N,\omega}\right\}$ for all $j\in \{1,\ldots, k\}\setminus~{\{j^{*}\}}$ and $\rho_{j^{*}}^{*} = \min\left\{\tfrac{n\mu^{2}+\theta^{2}}{n\mu^{2}+\theta^{2} + \sigma^{2}}, \rho_{N,\omega}\right\}$. With the same arguments as in the proof of Theorem~\ref{thm:expressivity-unrolling} we get that $\rho_{j^{*}}^{*}=\rho_{N,\omega}$ and after rearranging we get
  \begin{align}\label{eq:unrolling_iid_j_1_min_even}
   \cE_{i.i.d.}(\T) &=\tfrac{\mu^{2}}{2}n(\rho_{N,\omega}-1)^{2}\\ 
   &\quad + \tfrac{\theta^{2}}{2}\left[(k-1)\left(\min\left\{\tfrac{\theta^{2}}{\theta^{2} + \sigma^{2}}, \rho_{N,\omega}\right\}-1\right)^{2} + (n-k+1)(\rho_{N,\omega}-1)^{2}\right] \nonumber \\ 
   &\quad +  \tfrac{\sigma^{2}}{2}\left[(k-1)\min\left\{\tfrac{\theta^{2}}{\theta^{2} + \sigma^{2}}, \rho_{N,\omega}\right\}^{2} + (n-k+1)\rho_{N,\omega}^{2}\right]\nonumber \ . 
  \end{align}
  \item[$j^{*} \in \{k+1,\ldots,n\}$:] In that case it is 
  \begin{align}\label{eq:unrolling-iid-j-n}
   \cE_{i.i.d.}(\T) & =\tfrac{\mu^{2}}{2}n(\rho_{N,\omega}-1)^{2} + \tfrac{\theta^{2}}{2}\left(\sum_{j=1}^{k}(\rho_{j}-1)^{2} + (n-k)(\rho_{N,\omega}-1)^{2}\right)\\
     &\quad +  \tfrac{\sigma^{2}}{2}\left(\sum_{j=1}^{k}\rho_{j}^{2} + (n-k)\rho_{N,\omega}^{2}\right)\ . \nonumber
  \end{align}
  Optimizing over $\rho_{j}\leq \rho_{N,\omega}$ gives $\rho_{j}^{*} = \min\left\{\tfrac{\theta^{2}}{\theta^{2} + \sigma^{2}}, \rho_{N,\omega}\right\}$ for all $j \in \{1,\ldots, k\}$ with corresponding risk 
  \begin{align}\label{eq:unrolling_iid_j_n_min_even}
   \cE_{i.i.d.}(\T) &=\tfrac{\mu^{2}}{2}n(\rho_{N,\omega}-1)^{2}\\ 
   &\quad +\tfrac{\theta^{2}}{2}\left[k\left(\min\left\{\tfrac{\theta^{2}}{\theta^{2} + \sigma^{2}}, \rho_{N,\omega}\right\}-1\right)^{2} + (n-k)(\rho_{N,\omega}-1)^{2}\right]\nonumber \\
     &\quad + \tfrac{\sigma^{2}}{2}\left[k\min\left\{\tfrac{\theta^{2}}{\theta^{2} + \sigma^{2}}, \rho_{N,\omega}\right\}^{2} + (n-k)\rho_{N,\omega}^{2}\right]\ .  \nonumber
  \end{align}
   \end{description}
     Again, for $N$ even and $k<n$ the optimal risk is obtained for the minimum of (\ref{eq:unrolling_iid_j_1_min_even}) and (\ref{eq:unrolling_iid_j_n_min_even}). After rearrangement we get
  \begin{alignat*}{2}
    \cE_{i.i.d.}(\T) &= \tfrac{\mu^{2}}{2}n(\rho_{N,\omega}&& -1)^{2} + \tfrac{\theta^{2}}{2}(k-1)\left(\min\left\{\tfrac{\theta^{2}}{\theta^{2} + \sigma^{2}}, \rho_{N,\omega}\right\} - 1\right)^{2} + \tfrac{\theta^{2}}{2}(n-k)(\rho_{N,\omega}-1)^{2}\\ 
    &\quad + \tfrac{\sigma^{2}}{2}(k-&&1)\min\left\{\tfrac{\theta^{2}}{\theta^{2} + \sigma^{2}}, \rho_{N,\omega}\right\}^{2} + \tfrac{\sigma^{2}}{2}(n-k)\rho_{N,\omega}^{2}\\ 
    &\quad + \min\Big\{&&\tfrac{\theta^{2}}{2}(\rho_{N,\omega} -1)^{2} + \tfrac{\sigma^{2}}{2}\rho_{N,\omega}^{2},\\
    &\quad &&\tfrac{\theta^{2}}2\left(\min\left\{\tfrac{\theta^{2}}{\theta^{2} + \sigma^{2}}, \rho_{N,\omega}\right\} -1\right)^{2} + \tfrac{\sigma^{2}}2\min\left\{\tfrac{\theta^{2}}{\theta^{2} + \sigma^{2}}, \rho_{N,\omega}\right\}^{2} \Big\}
  \end{alignat*}
  which proves the claim for $N$ even and $k<n$.
  \item[$N$ odd:] For $N$ odd we have $\rho_{j} \geq c_{N,\omega}$. 
  \begin{description}
    \item[$j^{*} \in\{1,\ldots,k\}$:] In this case we also have to minimize $\cE_{i.i.d.}$ from (\ref{eq:unrolling_iid_j_1}) and some calculation shows that a minimization over $\rho_{j}\geq c_{N,\omega}$ gives $\rho_{j^{*}}^{*} = \max\left\{\tfrac{n\mu^{2}+\theta^{2}}{n\mu^{2}+\theta^{2} + \sigma^{2}}, c_{N,\omega}\right\}$ and $\rho_{j}^{*} = \max\left\{\tfrac{\theta^{2}}{\theta^{2} + \sigma^{2}}, c_{N,\omega}\right\}$ for all $j\in \{1,\ldots, k\}\setminus{\{j^{*}\}}$. The corresponding risk is 
    \begin{align}\label{eq:unrolling_iid_j_1_min_odd}
      \cE_{i.i.d.}(\T) &=\tfrac{\mu^{2}}{2}n\left(\max\left\{\tfrac{n\mu^{2}+\theta^{2}}{n\mu^{2}+\theta^{2} + \sigma^{2}}, c_{N,\omega}\right\}-1\right)^{2} + \tfrac{\theta^{2}}{2}\Big[\left(\max\left\{\tfrac{n\mu^{2}+\theta^{2}}{n\mu^{2}+\theta^{2} + \sigma^{2}}, c_{N,\omega}\right\}-1\right)^{2}\\ 
   &\quad + (k-1)\left(\max\left\{\tfrac{\theta^{2}}{\theta^{2} + \sigma^{2}}, c_{N,\omega}\right\}-1\right)^{2} + (n-k)(\rho_{N,\omega}-1)^{2}\Big]\nonumber\\ 
   &\quad +  \tfrac{\sigma^{2}}{2}\Big[\max\left\{\tfrac{n\mu^{2} + \theta^{2}}{n\mu^{2} + \theta^{2} + \sigma^{2}}, c_{N,\omega}\right\}^{2} + (k-1)\max\left\{\tfrac{\theta^{2}}{\theta^{2} + \sigma^{2}}, c_{N,\omega}\right\}^{2}\nonumber\\ 
   &\quad + (n-k)\rho_{N,\omega}^{2}\Big] \nonumber \ .
    \end{align}
    Note that this case only occurs if $(\rho_{j^{*}}^{*}-1)^{2}\leq (\rho_{N,\omega}-1)^{2}$.
    \item[$j^{*}\in\{k+1,\ldots,n\}$:] Optimizing (\ref{eq:unrolling-iid-j-n}) over $\rho_{j}\geq c_{N,\omega}$ gives $\rho_{j}^{*} = \max\left\{\tfrac{\theta^{2}}{\theta^{2} + \sigma^{2}}, c_{N,\omega}\right\}$ for all $j\in \{1,\ldots, k\}$ and thus 
    \begin{align}\label{eq:unrolling_iid_j_n_min_odd}
     \cE_{i.i.d.}(\T) &=\tfrac{\mu^{2}}{2}n(\rho_{N,\omega}-1)^{2} + \tfrac{\theta^{2}}{2}\Big(k(\max\{\tfrac{\theta^{2}}{\theta^{2} + \sigma^{2}}, c_{N,\omega}\}-1)^{2} + (n-k)(\rho_{N,\omega}-1)^{2}\Big)\\
     &\quad +  \tfrac{\sigma^{2}}{2}\Big(k\max\{\tfrac{\theta^{2}}{\theta^{2} + \sigma^{2}}, c_{N,\omega}\}^{2} + (n-k)\rho_{N,\omega}^{2}\Big).  \nonumber
    \end{align}
    This case occurs if $(\rho_{j}^{*}-1)^{2}\geq (\rho_{N,\omega}-1)^{2}$ for all $j\in\{1,\ldots,k\}$.
  \end{description}
  Therefore, for $N$ odd and $k<n$ the optimal risk is obtained for the minimum between (\ref{eq:unrolling_iid_j_1_min_odd}) and (\ref{eq:unrolling_iid_j_n_min_odd}). After rearrangement we get 
    \begin{alignat*}{2}
   \min_{\T\in\cB_{N,k,\omega}} \cE_{i.i.d.}(\T) &= \tfrac{\theta^{2}}{2}(k-1&&)\left(\max\left\{\tfrac{\theta^{2}}{\theta^{2} + \sigma^{2}}, c_{N,\omega}\right\} - 1\right)^{2} + \tfrac{\theta^{2}}{2}(n-k)(\rho_{N,\omega}-1)^{2}\\ 
   &\quad + \tfrac{\sigma^{2}}{2}(k-&&1)\max\left\{\tfrac{\theta^{2}}{\theta^{2} + \sigma^{2}}, c_{N,\omega}\right\}^{2} + \tfrac{\sigma^{2}}{2}(n-k)\rho_{N,\omega}^{2}\\ 
    &\quad + \min\Big\{&&\tfrac{\mu^{2}}{2}n\left(\max\left\{\tfrac{n\mu^{2}+\theta^{2}}{n\mu^{2}+\theta^{2} + \sigma^{2}}, c_{N,\omega}\right\} -1\right)^{2} + \tfrac{\sigma^{2}}{2} \max\left\{\tfrac{n\mu^{2}+\theta^{2}}{n\mu^{2}+\theta^{2} + \sigma^{2}}, c_{N,\omega}\right\}^{2}\\
    &\quad &&+ \tfrac{\theta^{2}}{2}\left(\max\left\{\tfrac{n\mu^{2}+\theta^{2}}{n\mu^{2}+\theta^{2} + \sigma^{2}}, c_{N,\omega}\right\} -1\right)^{2},\\
      &\quad &&\tfrac{\mu^{2}}{2}n(\rho_{N,\omega}-1)^{2} + \tfrac{\sigma^{2}}2\max\left\{\tfrac{\theta^{2}}{\theta^{2} + \sigma^{2}}, c_{N,\omega}\right\}^{2}\\ 
      &\quad &&+ \tfrac{\theta^{2}}2\left(\max\left\{\tfrac{\theta^{2}}{\theta^{2} + \sigma^{2}}, c_{N,\omega}\right\} -1\right)^{2} \Big\}
  \end{alignat*}
  which proves the claim for $N$ odd and $k<n$.
  \end{description}
It is left to show the result for $k=n$. In this case the risk is given by \begin{align*}
   \cE_{i.i.d.}(\T) & =\tfrac{\mu^{2}}{2}\sum_{j=1}^{n}\big(\rho_{j}-1\big)^{2}\abs{\scp{\vv_{j}}{\1}}^{2} + \tfrac{\theta^{2}}{2}\sum_{j=1}^{n}\big(\rho_{j}-1\big)^{2} + \tfrac{\sigma^{2}}{2}\sum_{j=1}^{n}\rho_{j}^{2} .
\end{align*} 
and no case distinction in $j^{*}$ is needed. Applying Lemma~\ref{lem:unrolling_orthonormal} with $c_{j} = (\rho_{j}-1)^{2}$, we get that 
\begin{align*}
  \cE_{i.i.d.}(\T) & =\tfrac{\mu^{2}}{2}\big(\rho_{j^{*}}-1\big)^{2}n + \tfrac{\theta^{2}}{2}\sum_{j=1}^{n}\big(\rho_{j}-1\big)^{2} + \tfrac{\sigma^{2}}{2}\sum_{j=1}^{n}\rho_{j}^{2} .
\end{align*} 
For $N$ even we optimize over $\rho_{j}\leq \rho_{N,\omega}$ and get $\rho_{j^{*}}^{*} = \min\left\{\tfrac{n\mu^{2}+\theta^{2}}{n\mu^{2}+\theta^{2}+\sigma^{2}},\rho_{N,\omega}\right\}$ and $\rho_{j}^{*} = \min\left\{\tfrac{\theta^{2}}{\theta^{2} + \sigma^{2}}, \rho_{N,\omega}\right\}$ for $j\neq j^{*}$. Thus, the optimal risk for $N$ even is 
\begin{align*}
  \cE_{i.i.d.}(\T) &= \tfrac{\mu^{2}}{2}n\left(\min\left\{\tfrac{n\mu^{2}+\theta^{2}}{n\mu^{2}+\theta^{2}+\sigma^{2}},\rho_{N,\omega}\right\}-1\right)^{2}\\
    &\quad + \tfrac{\theta^{2}}{2}\left[\left(\min\left\{\tfrac{n\mu^{2}+\theta^{2}}{n\mu^{2}+\theta^{2}+\sigma^{2}},\rho_{N,\omega}\right\}-1\right)^{2} + (n-1)\left(\min\left\{\tfrac{\theta^{2}}{\theta^{2} + \sigma^{2}}, \rho_{N,\omega}\right\}-1\right)^{2}\right]\\ 
    &\quad + \tfrac{\sigma^{2}}{2}\left[\min\left\{\tfrac{n\mu^{2}+\theta^{2}}{n\mu^{2}+\theta^{2}+\sigma^{2}},\rho_{N,\omega}\right\}^{2} + \tfrac{\sigma^{2}}{2}(n-1)\min\left\{\tfrac{\theta^{2}}{\theta^{2} + \sigma^{2}}, \rho_{N,\omega}\right\}^{2}\right].
\end{align*}
For $N$ odd, an optimization over $\rho_{j} \geq c_{N,\omega}$ results in $\rho_{j^{*}}^{*} = \max\left\{\tfrac{n\mu^{2}+\theta^{2}}{n\mu^{2}+\theta^{2}+\sigma^{2}},c_{N,\omega}\right\}$, $\rho_{j}^{*} = \max\left\{\tfrac{\theta^{2}}{\theta^{2} + \sigma^{2}},c_{N,\omega}\right\}$ for $j\neq j^{*}$ and 
\begin{align*}
     \cE_{i.i.d.}(\T) &= \tfrac{\mu^{2}}{2}n\left(\max\left\{\tfrac{n\mu^{2}+\theta^{2}}{n\mu^{2}+\theta^{2}+\sigma^{2}},c_{N,\omega}\right\}-1\right)^{2}\\
    &\quad + \tfrac{\theta^{2}}{2}\left[\left(\max\left\{\tfrac{n\mu^{2}+\theta^{2}}{n\mu^{2}+\theta^{2}+\sigma^{2}},c_{N,\omega}\right\}-1\right)^{2} + (n-1)\left(\max\left\{\tfrac{\theta^{2}}{\theta^{2} + \sigma^{2}}, c_{N,\omega}\right\}-1\right)^{2}\right]\\ 
    &\quad + \tfrac{\sigma^{2}}{2}\left[\max\left\{\tfrac{n\mu^{2}+\theta^{2}}{n\mu^{2}+\theta^{2}+\sigma^{2}},c_{N,\omega}\right\}^{2} + (n-1)\max\left\{\tfrac{\theta^{2}}{\theta^{2} + \sigma^{2}}, c_{N,\omega}\right\}^{2}\right] \ .
\end{align*}
which proves the remaining claim for $k=n$.
\end{proof}

\section*{Acknowledgments}
The authors thank Nicole Mücke for helpful remarks and valuable pointers to literature.

\bibliographystyle{plain}
\bibliography{refs}

\begin{thebibliography}{10}

\bibitem{tensorflow2015-whitepaper}
Mart\'{i}n Abadi, Ashish Agarwal, Paul Barham, et~al.
\newblock {TensorFlow}: Large-scale machine learning on heterogeneous systems,
  2015.
\newblock Software available from tensorflow.org.

\bibitem{ablin2019}
Pierre Ablin, Thomas Moreau, Mathurin Massias, and Alexandre Gramfort.
\newblock Learning step sizes for unfolded sparse coding.
\newblock {\em Advances in Neural Information Processing Systems}, 32, 2019.

\bibitem{adler2018}
Jonas Adler and Ozan {\"O}ktem.
\newblock Learned primal-dual reconstruction.
\newblock {\em IEEE transactions on medical imaging}, 37(6):1322--1332, 2018.

\bibitem{adler2017}
Jonas Adler, Axel Ringh, Ozan {\"O}ktem, and Johan Karlsson.
\newblock Learning to solve inverse problems using wasserstein loss.
\newblock {\em arXiv preprint arXiv:1710.10898}, 2017.

\bibitem{aggarwal2018}
Hemant~K Aggarwal, Merry~P Mani, and Mathews Jacob.
\newblock Modl: Model-based deep learning architecture for inverse problems.
\newblock {\em IEEE transactions on medical imaging}, 38(2):394--405, 2018.

\bibitem{alberti2021learning}
Giovanni~S Alberti, Ernesto De~Vito, Matti Lassas, Luca Ratti, and Matteo
  Santacesaria.
\newblock Learning the optimal tikhonov regularizer for inverse problems.
\newblock {\em Advances in Neural Information Processing Systems}, 34, 2021.

\bibitem{arridge2019}
Simon Arridge, Peter Maass, Ozan {\"O}ktem, and Carola-Bibiane Sch{\"o}nlieb.
\newblock Solving inverse problems using data-driven models.
\newblock {\em Acta Numerica}, 28:1--174, 2019.

\bibitem{bard2013}
Jonathan~F Bard.
\newblock {\em Practical bilevel optimization: algorithms and applications},
  volume~30.
\newblock Springer Science \& Business Media, 2013.

\bibitem{beck2009}
Amir Beck and Marc Teboulle.
\newblock A fast iterative shrinkage-thresholding algorithm with application to
  wavelet-based image deblurring.
\newblock In {\em 2009 IEEE International Conference on Acoustics, Speech and
  Signal Processing}, pages 693--696. IEEE, 2009.

\bibitem{bertocchi2020}
Carla Bertocchi, Emilie Chouzenoux, Marie-Caroline Corbineau, Jean-Christophe
  Pesquet, and Marco Prato.
\newblock Deep unfolding of a proximal interior point method for image
  restoration.
\newblock {\em Inverse Problems}, 36(3):034005, 2020.

\bibitem{Blekherman:Parrilo:Thomas}
G.~Blekherman, P.A. Parrilo, and R.R. Thomas.
\newblock {\em Semidefinite Optimization and Convex Algebraic Geometry},
  volume~13 of {\em MOS-SIAM Series on Optimization}.
\newblock SIAM and the Mathematical Optimization Society, Philadelphia, 2013.

\bibitem{bonettini2009}
Silvia Bonettini and Thomas Serafini.
\newblock Non-negatively constrained image deblurring with an inexact interior
  point method.
\newblock {\em Journal of Computational and Applied Mathematics},
  231(1):236--248, 2009.

\bibitem{boykov2001}
Yuri Boykov, Olga Veksler, and Ramin Zabih.
\newblock Fast approximate energy minimization via graph cuts.
\newblock {\em IEEE Transactions on pattern analysis and machine intelligence},
  23(11):1222--1239, 2001.

\bibitem{bracken1973}
Jerome Bracken and James~T McGill.
\newblock Mathematical programs with optimization problems in the constraints.
\newblock {\em Operations Research}, 21(1):37--44, 1973.

\bibitem{brauer2016}
Christoph Brauer, Timo Gerkmann, and Dirk Lorenz.
\newblock Sparse reconstruction of quantized speech signals.
\newblock In {\em 2016 IEEE International Conference on Acoustics, Speech and
  Signal Processing (ICASSP)}, pages 5940--5944. IEEE, 2016.

\bibitem{brauer2019}
Christoph Brauer, Ziyue Zhao, Dirk Lorenz, and Tim Fingscheidt.
\newblock Learning to dequantize speech signals by primal-dual networks: an
  approach for acoustic sensor networks.
\newblock In {\em ICASSP 2019-2019 IEEE International Conference on Acoustics,
  Speech and Signal Processing (ICASSP)}, pages 7000--7004. IEEE, 2019.

\bibitem{chambolle2011}
Antonin Chambolle and Thomas Pock.
\newblock A first-order primal-dual algorithm for convex problems with
  applications to imaging.
\newblock {\em Journal of mathematical imaging and vision}, 40(1):120--145,
  2011.

\bibitem{chen2021b}
Tianlong Chen, Xiaohan Chen, Wuyang Chen, Howard Heaton, Jialin Liu, Zhangyang
  Wang, and Wotao Yin.
\newblock Learning to optimize: A primer and a benchmark.
\newblock {\em arXiv preprint arXiv:2103.12828}, 2021.

\bibitem{chen2018}
Xiaohan Chen, Jialin Liu, Zhangyang Wang, and Wotao Yin.
\newblock Theoretical linear convergence of unfolded ista and its practical
  weights and thresholds.
\newblock {\em Advances in Neural Information Processing Systems}, 31, 2018.

\bibitem{Chen2014}
Yunjin Chen, Wensen Feng, Ren{\'e} Ranftl, Hong Qiao, and Thomas Pock.
\newblock A higher-order mrf based variational model for multiplicative noise
  reduction.
\newblock {\em IEEE signal processing letters}, 21(11):1370--1374, 2014.

\bibitem{colson2007}
Beno{\^\i}t Colson, Patrice Marcotte, and Gilles Savard.
\newblock An overview of bilevel optimization.
\newblock {\em Annals of operations research}, 153(1):235--256, 2007.

\bibitem{CuZh2007}
Felipe Cucker and Ding~Xuan Zhou.
\newblock {\em Learning theory: an approximation theory viewpoint}, volume~24.
\newblock Cambridge University Press, 2007.

\bibitem{cuturi2013}
Marco Cuturi.
\newblock Sinkhorn distances: Lightspeed computation of optimal transport.
\newblock {\em Advances in neural information processing systems}, 26, 2013.

\bibitem{daubechies2004}
Ingrid Daubechies, Michel Defrise, and Christine De~Mol.
\newblock An iterative thresholding algorithm for linear inverse problems with
  a sparsity constraint.
\newblock {\em Communications on Pure and Applied Mathematics: A Journal Issued
  by the Courant Institute of Mathematical Sciences}, 57(11):1413--1457, 2004.

\bibitem{dempe2020}
Stephan Dempe and Alain Zemkoho.
\newblock Bilevel optimization.
\newblock In {\em Springer optimization and its applications. Vol. 161}.
  Springer, 2020.

\bibitem{dietterich1995}
Tom Dietterich.
\newblock Overfitting and undercomputing in machine learning.
\newblock {\em ACM computing surveys (CSUR)}, 27(3):326--327, 1995.

\bibitem{dobriban2018high}
Edgar Dobriban and Stefan Wager.
\newblock High-dimensional asymptotics of prediction: Ridge regression and
  classification.
\newblock {\em The Annals of Statistics}, 46(1):247--279, 2018.

\bibitem{engl1996regularization}
Heinz~Werner Engl, Martin Hanke, and Andreas Neubauer.
\newblock {\em Regularization of inverse problems}, volume 375.
\newblock Springer Science \& Business Media, 1996.

\bibitem{ghadimi2018}
Saeed Ghadimi and Mengdi Wang.
\newblock Approximation methods for bilevel programming.
\newblock {\em arXiv preprint arXiv:1802.02246}, 2018.

\bibitem{goodfellow2016}
Ian Goodfellow, Yoshua Bengio, and Aaron Courville.
\newblock {\em Deep learning}.
\newblock MIT press, 2016.

\bibitem{gregor2010}
Karol Gregor and Yann LeCun.
\newblock Learning fast approximations of sparse coding.
\newblock In {\em Proceedings of the 27th international conference on
  international conference on machine learning}, pages 399--406, 2010.

\bibitem{Hastie2009}
Trevor Hastie, Robert Tibshirani, and Jerome Friedman.
\newblock {\em Overview of Supervised Learning}, pages 9--41.
\newblock Springer New York, New York, NY, 2009.

\bibitem{Iliman:deWolff:Circuits}
S.~Iliman and T.~de~Wolff.
\newblock Amoebas, nonnegative polynomials and sums of squares supported on
  circuits.
\newblock {\em Res. Math. Sci.}, 3(9), 2016.

\bibitem{Iliman:deWolff:GP}
S.~Iliman and T.~de~Wolff.
\newblock Lower bounds for polynomials with simplex newton polytopes based on
  geometric programming.
\newblock {\em SIAM J. Optim.}, 26(2):1128--1146, 2016.

\bibitem{jin2012}
Bangti Jin and Peter Maass.
\newblock Sparsity regularization for parameter identification problems.
\newblock {\em Inverse Problems}, 28(12):123001, 2012.

\bibitem{kaipio2006statistical}
Jari Kaipio and Erkki Somersalo.
\newblock {\em Statistical and computational inverse problems}, volume 160.
\newblock Springer Science \& Business Media, 2006.

\bibitem{kappes2013}
Joerg Kappes, Bjoern Andres, Fred Hamprecht, Christoph Schnorr, Sebastian
  Nowozin, Dhruv Batra, Sungwoong Kim, Bernhard Kausler, Jan Lellmann, Nikos
  Komodakis, et~al.
\newblock A comparative study of modern inference techniques for discrete
  energy minimization problems.
\newblock In {\em Proceedings of the IEEE conference on computer vision and
  pattern recognition}, pages 1328--1335. IEEE, 2013.

\bibitem{kingma2014adam}
Diederik~P Kingma and Jimmy Ba.
\newblock Adam: A method for stochastic optimization.
\newblock {\em arXiv preprint arXiv:1412.6980}, 2014.

\bibitem{kobler2020total}
Erich Kobler, Alexander Effland, Karl Kunisch, and Thomas Pock.
\newblock Total deep variation for linear inverse problems.
\newblock In {\em Proceedings of the IEEE/CVF Conference on computer vision and
  pattern recognition}, pages 7549--7558, 2020.

\bibitem{kolmogorov2006}
Vladimir Kolmogorov and Carsten Rother.
\newblock Comparison of energy minimization algorithms for highly connected
  graphs.
\newblock In {\em European Conference on Computer Vision}, pages 1--15.
  Springer, 2006.

\bibitem{landweber1951iteration}
Louis Landweber.
\newblock An iteration formula for fredholm integral equations of the first
  kind.
\newblock {\em American journal of mathematics}, 73(3):615--624, 1951.

\bibitem{Lasserre:IntroductionPolynomialandSemiAlgebraicOptimization}
J.B. Lasserre.
\newblock {\em An Introduction to Polynomial and Semi-Algebraic Optimization},
  volume~1 of {\em Cambridge Texts in Applied Mathematics}.
\newblock Cambridge University Press, 2015.

\bibitem{le2022faster}
Hoang Trieu~Vy Le, Nelly Pustelnik, and Marion Foare.
\newblock The faster proximal algorithm, the better unfolded deep learning
  architecture? the study case of image denoising.
\newblock In {\em 2022 30th European Signal Processing Conference (EUSIPCO)},
  pages 947--951. IEEE, 2022.

\bibitem{lecun2015}
Yann LeCun, Yoshua Bengio, and Geoffrey Hinton.
\newblock Deep learning.
\newblock {\em nature}, 521(7553):436--444, 2015.

\bibitem{li2009}
Yingying Li and Stanley Osher.
\newblock Coordinate descent optimization for l 1 minimization with application
  to compressed sensing; a greedy algorithm.
\newblock {\em Inverse Problems \& Imaging}, 3(3):487, 2009.

\bibitem{liu2019a}
Jialin Liu and Xiaohan Chen.
\newblock Alista: Analytic weights are as good as learned weights in lista.
\newblock In {\em International Conference on Learning Representations (ICLR)},
  2019.

\bibitem{liu2019}
Risheng Liu, Shichao Cheng, Long Ma, Xin Fan, and Zhongxuan Luo.
\newblock Deep proximal unrolling: Algorithmic framework, convergence analysis
  and applications.
\newblock {\em IEEE Transactions on Image Processing}, 28(10):5013--5026, 2019.

\bibitem{loizou2013speech}
Philipos~C Loizou.
\newblock {\em Speech enhancement: theory and practice}.
\newblock CRC press, 2013.

\bibitem{malezieux2022}
Beno{\^\i}t Mal{\'e}zieux, Thomas Moreau, and Matthieu Kowalski.
\newblock Understanding approximate and unrolled dictionary learning for
  pattern recovery.
\newblock In {\em International Conference on Learning Representations}, 2022.

\bibitem{monga2021}
Vishal Monga, Yuelong Li, and Yonina~C. Eldar.
\newblock Algorithm unrolling: Interpretable, efficient deep learning for
  signal and image processing.
\newblock {\em IEEE Signal Processing Magazine}, 38(2):18--44, 2021.

\bibitem{mucke2022data}
Nicole M{\"u}cke, Enrico Reiss, Jonas Rungenhagen, and Markus Klein.
\newblock Data-splitting improves statistical performance in overparameterized
  regimes.
\newblock In {\em International Conference on Artificial Intelligence and
  Statistics}, pages 10322--10350. PMLR, 2022.

\bibitem{nikolova2011}
Mila Nikolova.
\newblock Energy minimization methods.
\newblock In Otmar Scherzer, editor, {\em Handbook of Mathematical Methods in
  Imaging}, pages 138--186. Springer, 2011.

\bibitem{olshausen1996}
Bruno~A Olshausen and David~J Field.
\newblock Emergence of simple-cell receptive field properties by learning a
  sparse code for natural images.
\newblock {\em Nature}, 381(6583):607--609, 1996.

\bibitem{ongie2020}
Gregory Ongie, Ajil Jalal, Christopher~A Metzler, Richard~G Baraniuk,
  Alexandros~G Dimakis, and Rebecca Willett.
\newblock Deep learning techniques for inverse problems in imaging.
\newblock {\em IEEE Journal on Selected Areas in Information Theory},
  1(1):39--56, 2020.

\bibitem{roth2005fields}
Stefan Roth and Michael~J Black.
\newblock Fields of experts: A framework for learning image priors.
\newblock In {\em 2005 IEEE Computer Society Conference on Computer Vision and
  Pattern Recognition (CVPR'05)}, volume~2, pages 860--867. IEEE, 2005.

\bibitem{Roth2009}
Stefan Roth and Michael~J Black.
\newblock Fields of experts.
\newblock {\em International Journal of Computer Vision}, 82(2):205--229, 2009.

\bibitem{samek2019explainable}
Wojciech Samek, Gr{\'e}goire Montavon, Andrea Vedaldi, Lars~Kai Hansen, and
  Klaus-Robert M{\"u}ller.
\newblock {\em Explainable AI: interpreting, explaining and visualizing deep
  learning}, volume 11700.
\newblock Springer Nature, 2019.

\bibitem{scherzer2009}
Otmar Scherzer, Markus Grasmair, Harald Grossauer, Markus Haltmeier, and Frank
  Lenzen.
\newblock {\em Variational Methods in Imaging}.
\newblock Springer New York, NY, 2009.

\bibitem{sinha2017}
Ankur Sinha, Pekka Malo, and Kalyanmoy Deb.
\newblock A review on bilevel optimization: from classical to evolutionary
  approaches and applications.
\newblock {\em IEEE Transactions on Evolutionary Computation}, 22(2):276--295,
  2017.

\bibitem{szeliski2006}
Richard Szeliski, Ramin Zabih, Daniel Scharstein, Olga Veksler, Vladimir
  Kolmogorov, Aseem Agarwala, Marshall Tappen, and Carsten Rother.
\newblock A comparative study of energy minimization methods for markov random
  fields.
\newblock In {\em European conference on computer vision}, pages 16--29.
  Springer, 2006.

\bibitem{takabe2020}
Satoshi Takabe and Tadashi Wadayama.
\newblock Theoretical interpretation of learned step size in deep-unfolded
  gradient descent.
\newblock {\em arXiv preprint arXiv:2001.05142}, 2020.

\bibitem{tian2020deep}
Chunwei Tian, Lunke Fei, Wenxian Zheng, Yong Xu, Wangmeng Zuo, and Chia-Wen
  Lin.
\newblock Deep learning on image denoising: An overview.
\newblock {\em Neural Networks}, 131:251--275, 2020.

\bibitem{vese2016}
Luminita~A Vese and Carole Le~Guyader.
\newblock {\em Variational methods in image processing}.
\newblock CRC Press Boca Raton, FL, 2016.

\bibitem{vu2021}
Huy Vu, Gene Cheung, and Yonina~C Eldar.
\newblock Unrolling of deep graph total variation for image denoising.
\newblock In {\em ICASSP 2021-2021 IEEE International Conference on Acoustics,
  Speech and Signal Processing (ICASSP)}, pages 2050--2054. IEEE, 2021.

\bibitem{zheng2021deep}
Hongyi Zheng, Hongwei Yong, and Lei Zhang.
\newblock Deep convolutional dictionary learning for image denoising.
\newblock In {\em Proceedings of the IEEE/CVF Conference on Computer Vision and
  Pattern Recognition}, pages 630--641, 2021.

\end{thebibliography}

\end{document}